\documentclass[fleqn,10pt]{wlscirep}
\usepackage[utf8]{inputenc}
\usepackage[T1]{fontenc}
\usepackage{lineno}
\usepackage{tikz}
\usepackage{todonotes}
\usepackage{pgfplots}
\usepackage{array} 
\usepackage{svg}
\usepackage{graphicx}
\usepackage{nameref}
\usepackage{hyperref} 
\usepackage{amssymb} 
\usepackage{amsmath}
\usepackage{pifont}
\newcommand{\cmark}{\ding{51}} 
\newcommand{\xmark}{\ding{55}} 
\usepackage{enumitem}

\usepackage{caption}
\usepackage{subcaption}

\usepackage[acronym,]{glossaries}
\newacronym{hmm}{HMM}{Human Muscular Manipulability}
\newacronym{kmi}{KMI}{Kinematic Manipulability Index}
\newacronym{dmi}{DMI}{Dynamic Manipulability Index}
\newacronym{lci}{LCI}{Local Conditionint Index}
\newacronym{emg}{EMG}{Electromyography}
\newacronym{mae}{MAE}{Mean Absolute Error}
\newacronym{ae}{AE}{Absolute Error}
\newacronym{imu}{IMU}{Inertial Measurement Unit}

\usepackage{array}

\newenvironment{conditions}
  {\par\vspace{\abovedisplayskip}\noindent\begin{tabular}{>{$}l<{$} @{${}={}$} l}}
  {\end{tabular}\par\vspace{\belowdisplayskip}}

\title{CHIRLA: Comprehensive High-resolution Identification and Re-identification for Large-scale Analysis}

\author[1*]{Bessie Dominguez-Dager}
\author[1]{Felix Escalona}
\author[1]{Francisco Gomez-Donoso}
\author[1]{Miguel Cazorla}

\affil[1]{Institute for Computing Research, P.O. Box 99. 03080, Alicante, Spain.}
\affil[*]{corresponding author(s): Bessie Dominguez-Dager (bessie.dominguez@ua.es)}

\begin{abstract}

Person re-identification (Re-ID) is a key challenge in computer vision, requiring the matching of individuals across cameras, locations, and time. While most research focuses on short-term scenarios with minimal appearance changes, real-world applications demand robust systems that handle long-term variations caused by clothing and physical changes.
We present CHIRLA, Comprehensive High-resolution Identification and Re-identification for Large-scale Analysis, a novel dataset designed for video-based long-term person Re-ID. 
CHIRLA was recorded over seven months in four connected indoor environments using seven strategically placed cameras, capturing realistic movements with substantial clothing and appearance variability. The dataset includes 22 individuals, more than five hours of video, and about 1M bounding boxes with identity annotations obtained through semi-automatic labeling. We also define benchmark protocols for person tracking and Re-ID, covering diverse and challenging scenarios such as occlusion, reappearance, and multi-camera conditions. By introducing this comprehensive benchmark, we aim to facilitate the development and evaluation of Re-ID algorithms that can reliably perform in challenging, long-term real-world scenarios. The benchmark code is publicly available at: \url{https://github.com/bdager/CHIRLA}.

\end{abstract}
\pgfplotsset{compat=1.18} 
\begin{document}

\flushbottom
\maketitle

\section*{Background \& Summary}


{ {

Person re-identification (Re-ID) is a challenging problem in computer vision that involves matching individuals across different camera views, locations and times. This task becomes even more complex when considering long-term scenarios, where individuals may change their appearance due to variations in clothing or alterations in physical characteristics. 

Early Re-ID benchmarks such as ETH \cite{schwartz09d}, GRID \cite{loy2010time}, VIPeR \cite{gray2008viewpoint}, and CAVIAR \cite{cheng2011custom} focused on short-term conditions, typically with low-resolution images and limited temporal information. With the advent of deep learning, larger datasets such as CUHK03 \cite{li2014deepreid}, RAiD \cite{das2014consistent}, Market-1501 \cite{zheng2015scalable}, PRW \cite{zheng2017person}, DukeMTMC \cite{ristani2016performance} and MSMT17  \cite{wei2018person} expanded scale and camera coverage, offering more identities (IDs) and additional views of the same person across cameras. 
However, these datasets remain fundamentally short-term: once an individual appears in one sequence, they are rarely annotated across different sessions, which limits their utility for long-term evaluation.

Video-based datasets such as MARS \cite{zheng2016mars} advanced the field by providing video sequences with bounding box annotations, enabling temporal modeling and integration with tracking methods. Nevertheless, it remains short-term, restricted to recordings of only a few seconds per ID, and was not designed to capture deliberate appearance variation such as clothing changes.

Clothing-change benchmarks were later introduced to capture long-term variability. Motion-ReID \cite{zhang2018long} emerged as a video-based person Re-ID dataset, but it is constrained to very limited appearance changes, providing only two different outfits per ID recorded from two indoor disjoint static surveillance cameras.

Other datasets attempted to address clothing-change variability, but they are mainly image-based.
Celeb-reID \cite{huang2019celebrities} consists of images collected from Internet street snapshots of celebrities. It is valuable for studying clothing variation and provides diverse appearance changes. However, the images are sourced from photographers’ snapshots, which do not replicate the camera angles, viewpoints, or environmental conditions typical of surveillance systems. PRCC \cite{yang2019person} contains individuals captured by three cameras under two outfits per individual. Its paired design makes it suitable for controlled clothing-change evaluation, but it is limited in the extent of appearance variation, constrained to a small number of viewpoints, and remains image-based. 
LTCC \cite{qian2020long} expands the setting by including more cameras and recording across multiple days with clothing labels, thereby increasing viewpoint coverage. Yet, like Celeb-reID and PRCC, LTCC consists of independent images without temporal continuity.

DeepChange \cite{xu2023deepchange} represents the largest dataset targeting long-term Re-ID under clothing change to date, collected over 12 months from 17 outdoor surveillance cameras. It provides valuable large-scale coverage of appearance variation. 
However, while the data originates from real-world video recordings, the public release consists only of low-resolution, fixed-size cropped bounding-box images rather than continuous tracklets or videos. This format, while realistic for outdoor surveillance, restricts evaluation primarily to body-based features and makes it difficult to assess Re-ID methods that also incorporate facial cues, which are particularly relevant in indoor scenarios. It also limits the dataset’s direct use of methods that can combine and take advantage of tracking or occlusion-aware benchmarks.

Table \ref{tab:datasetreview} summarizes the characteristics of the datasets discussed above. 
Overall, to the best of our knowledge, there is no existing person Re-ID dataset that is both long-term and video-based, offering temporal continuity, clothing variation, and image quality suitable for indoor surveillance analysis.

\begin{table}[!htb]
    \centering
    {
    \begin{tabular}{|l|l|l|l|l|l|l|l|l|l|}
    \hline
    Datasets & 
    \begin{tabular}[l]{@{}c@{}}\# \\ Videos\end{tabular} & 
    \begin{tabular}[l]{@{}c@{}}Avg. Time\\ /Video (s)\end{tabular} & 
    \# IDs & 
    \# Bboxes &
    \begin{tabular}[l]{@{}c@{}} Avg.\\ \#Bboxes/ID \end{tabular} &
    \begin{tabular}[l]{@{}c@{}}Image\\ Resolution\end{tabular}  & 
    \begin{tabular}[l]{@{}c@{}} \# \\ Cam.\end{tabular} & 
    { CC} &
    TC 
    \\
    \hline
    ETH \cite{schwartz09d}          & 4  & 38  & 146 & 8580   & 59  & 640×480   & 1  & \xmark & \xmark \\ \hline
    GRID \cite{loy2010time}         & -  & -   & 250 & 1,275  & 5   & 320×230   & 8  & \xmark & \xmark \\ \hline
    VIPeR \cite{gray2008viewpoint}  & -  & -   & 632 & 1,264  & 2   & 128×48    & 2  & \xmark & \xmark  \\ \hline
    CAVIAR \cite{cheng2011custom}   & -  & -   & 72  & 1,220  & 17  & 72×144    & 2  & \xmark & \xmark \\ \hline
    CUHK03 \cite{li2014deepreid}    & -  & -   & 1,360 & 13,164 & 10 & Variable  & 6  & \xmark & \xmark \\ \hline
    RAiD \cite{das2014consistent}   & -  & -   & 43  & 6,920  & 161 & 128×64    & 4  & \xmark & \xmark \\ \hline
    Market1501 \cite{zheng2015scalable} & - & - & 1,501 & 32,668 & 22 & 1080×1920 & 6  & \xmark & \xmark \\ \hline
    PRW \cite{zheng2017person}      & - & - & 932 & 34,304 & 37 & 1080×1920 & 6 & \xmark & \xmark \\ \hline 
    DukeMTMC \cite{ristani2016performance} & - & - & 1,812 & 36,411 & 20 & 1920×1080 & 8 & \xmark & \xmark \\ \hline
    MSMT17 \cite{wei2018person}     & -  & -   & 4,101 & 126,441 & 31 & Variable  & 15 & \xmark & \xmark \\ \hline
    MARS \cite{zheng2016mars}       & 20,715 & 2 & 1,261 & 1,067,516 & 847 & 1080×1920 & 6 & \xmark & \cmark \\ \hline       
    Motion-ReID \cite{zhang2018long} & 240 & 3.3 & 30 & 24,480 & 816 & Variable & 2 & \cmark & \cmark \\ \hline
    Celeb-reID \cite{huang2019celebrities} & - & - & 590 & 10,842 & 18 & 128×64 & 1 & \cmark & \xmark \\ \hline
    PRCC \cite{yang2019person}      & - & - & 221 & 33,698 & 152 & Variable & 3 & \cmark & \xmark \\ \hline  
    LTCC \cite{qian2020long}        & - & - & 152 & 17,138 & 113 & Variable & 12 & \cmark & \xmark \\ \hline
    DeepChange \cite{xu2023deepchange} & - & - & 1,121 & 178,407 & 159 & 1920×1080 & 17 & \cmark & \xmark \\ \hline
    \textbf{CHIRLA} & 70 & 284 & 22 & 963,554 & \textbf{43,798} & 1080×720 & 7 & \cmark & \cmark \\ \hline

    \end{tabular}}
    \caption{{ Summary of the reviewed datasets and their main characteristics. CC: clothing-change; TC: temporal coherence, indicating that cameras recorded simultaneously and were synchronized across views; and the recorded videos are provided to allow analysis of cross-camera person Re-ID. Image resolution corresponds to the maximum recording resolution available in each dataset; if multiple camera resolutions are used, only the highest is reported. For datasets that provide only cropped bounding boxes, the bounding-box resolution is reported. When marked as Variable, crops are either automatically detected by a person detector or annotated without fixed size constraints.}}
    \label{tab:datasetreview}    
\end{table}






To address these gaps, we introduce CHIRLA, a novel dataset for long-term person Re-ID in indoor surveillance environments. 
CHIRLA was recorded over seven months (May–December) in a multi-room office space, capturing realistic movement patterns across hallways, shared workspaces, and entry points. Recordings were obtained from multiple cameras strategically placed around the office, covering various angles and perspectives, often ensuring visible faces in addition to body appearance.
With this setup, CHIRLA includes variable resolutions depending on camera placement: lower-resolution crops in distant hallway views and higher-resolution crops in closer office settings. 
To further enhance the complexity of the dataset, participants were instructed to change their clothing between sessions, introducing significant appearance variability over time.

CHIRLA dataset introduces several unique features:
(1) More bounding boxes ($\sim964K$) and more samples per ID ($\sim44K$) than previous clothing-change datasets, enabling deeper temporal analysis of the same individuals.
(2) Continuous video sequences with detailed annotations, explicitly capturing occlusions (partial and complete), reappearances, and multi-camera transitions.
(3) Fine-grained benchmark protocols for both Re-ID and tracking under occlusion, reappearance, and viewpoint changes.
(4) Open-set evaluation with distractors: known IDs appear in both training and testing, while unknown IDs (distractors) appear only in testing, forcing models to set realistic decision thresholds.


In summary, CHIRLA complements existing datasets by providing temporally grounded and scenario-specific evaluations, while also enabling face-aware analysis thanks to the higher resolution of its person bounding boxes. Although it contains fewer IDs than large-scale outdoor benchmarks, CHIRLA offers richer per-ID data, an indoor surveillance setup, and realistic protocols for long-term Re-ID and tracking applications.
CHIRLA is not designed as a large-scale training dataset, but rather as a complementary benchmark for evaluating the robustness of models trained on larger datasets under realistic long-term and indoor surveillance conditions.

We make the following contributions: 
\begin{enumerate}
    \item We introduce CHIRLA, the first fully annotated, long-term, video-based person Re-ID dataset in an indoor environment, featuring challenging clothing changes, multi-camera views, and higher-resolution bounding boxes than existing datasets. CHIRLA also provides more bounding boxes and more samples per ID than previous clothing-change datasets. 
    \item We define specific benchmark protocols, not only for person Re-ID but also for person tracking, covering diverse and challenging scenarios such as occlusion, reappearance, and multi-camera transitions.
    \item We conduct extensive experiments with CNN-based models \cite{He_2016_CVPR, pan2018IBN-Net, Liu_2022_CVPR, maaz2022edgenext, ghostnet}, Vision Transformers \cite{dosovitskiy2020vit, liu2021swin, yuan2022volo, vasu2023fastvit, wang2024repvit}, and state-of-the-art tracking methods  \cite{zhang2022bytetrack, aharon2022botsort, cao2023ocsort, maggiolino2023deepocsort, du2023strongsort, stanojevic2024boosttrack, stanojevic2024btpp}, demonstrating that CHIRLA remains highly challenging and serves as a valuable benchmark for future research.  
\end{enumerate}



}
}

\section*{Methods}
\label{sec:Methods}


In this section, we describe the capture setup and environment, including the specifications and placement of the cameras, the interconnection architecture, and the software used for data capture, organization, and labeling.

This study was conducted under the supervision of the Ethical Committee of the University of Alicante (Approval No. UA-2022-11-12). All participants were informed about the study, including data handling procedures, and provided their consent for participation and unrestricted data use.  Specifically, participants granted permission for their inclusion in dataset video recordings, which capture physical data (face and body features), and consented to making these recordings publicly available with unrestricted use.
The dataset has been made publicly available under the CC-BY license.




\subsection*{Capture Setup}
\label{subsec:capture-setup}


The dataset was recorded in the Robotics, Vision and Intelligent Systems Research Group headquarters at the University of Alicante, Spain, which displays a typical office environment. The capture setup involves two main items: the physical spaces and the cameras placed in them. 

\begin{figure}[!htb]
    \centering
    \includegraphics[width=0.5\linewidth]{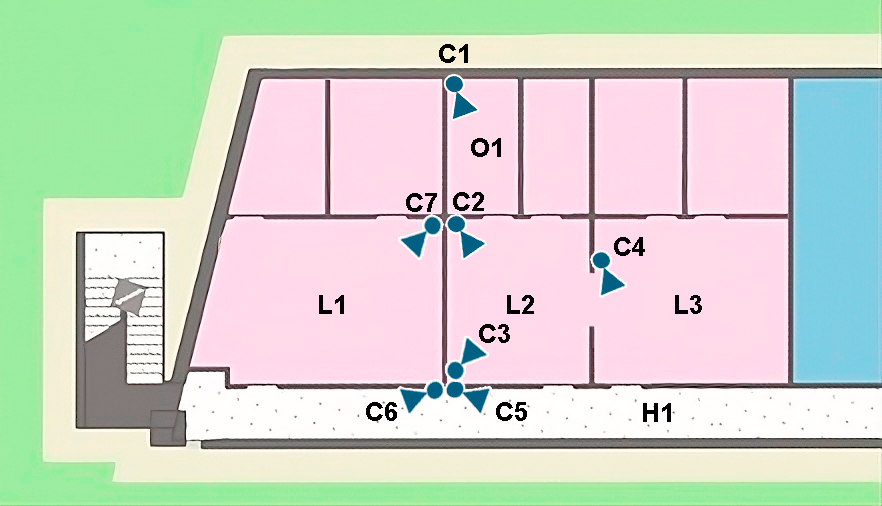}
    \caption{Position and orientation of the cameras within the environment}
    \label{fig:camera_position}
\end{figure}

Regarding the physical spaces in which the dataset has been captured, the layout of the area covered by the cameras is shown in Figure \ref{fig:camera_position}. The environment features a typical office setting with computers, desks, chairs, panels and whiteboards. Some objects eventually appear in some videos such as robots, cardboard boxes or others. As the videos are captured during a large time frame, the appearance of the environments slightly changed from one to another. There are three large working areas, one large corridor and one small office. These areas have the following settings. L1 is a big laboratory of 33.3 $m^2$ of area that is connected to the main corridor H1 and has a camera C7 placed in the opposite corner of the door, pointing at it. L2 is another laboratory of 20.64 $m^2$ that contains two cameras: C2, which is located in the opposite corner of the door, pointing at it; and C3, which is located in the south-west corner of the room, pointing at the office O1 door. This room is connected to another laboratory L3 and the corridor H1. The last laboratory is L3, which features 27.93 $m^2$ and has one camera C4 placed above the door that connects L2 and L3, and is oriented to the door. L3 is connected to the corridor H1 and the laboratory L2. There is also an individual office O1, that has 8,48 $m^2$ of area. It has one camera C1 located in the opposite corner of the door, oriented at it. The office is connected to the laboratory L2. Finally, the corridor H1 is also involved. It has an area of 44.01 $m^2$ that is covered by camera C6 and C5. H1 is connected to the three laboratories L1, L2 and L3.

\begin{figure}[!htb]
    \centering
    \includegraphics[width=0.24\linewidth, height=2.5cm]{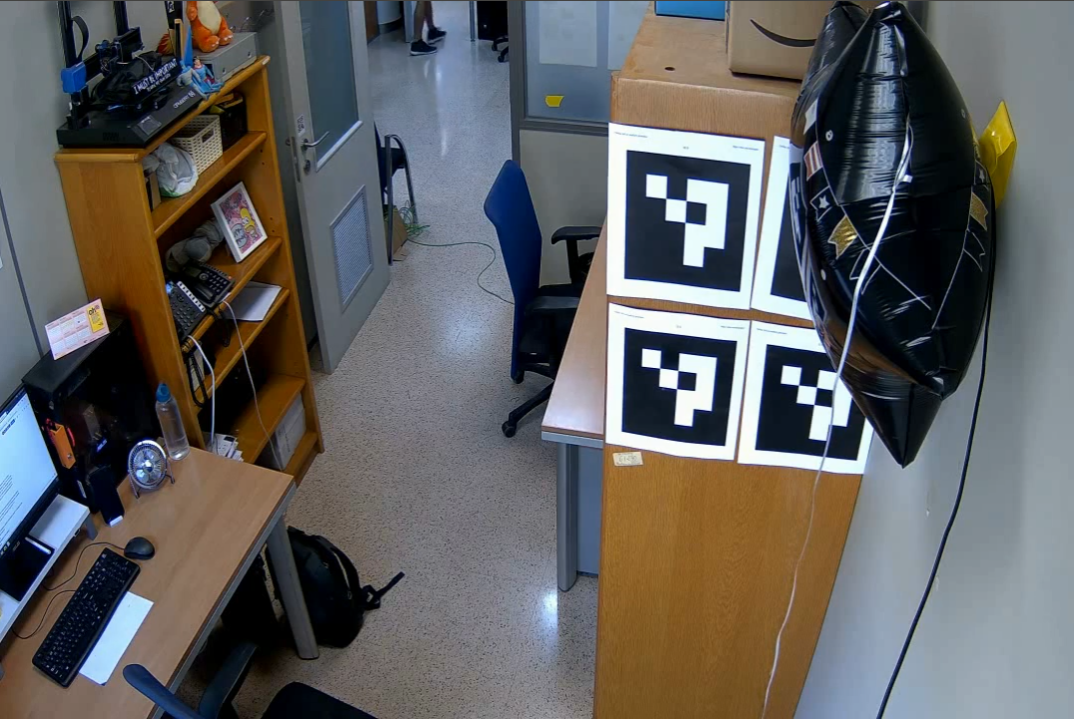}
    \includegraphics[width=0.24\linewidth, height=2.5cm]{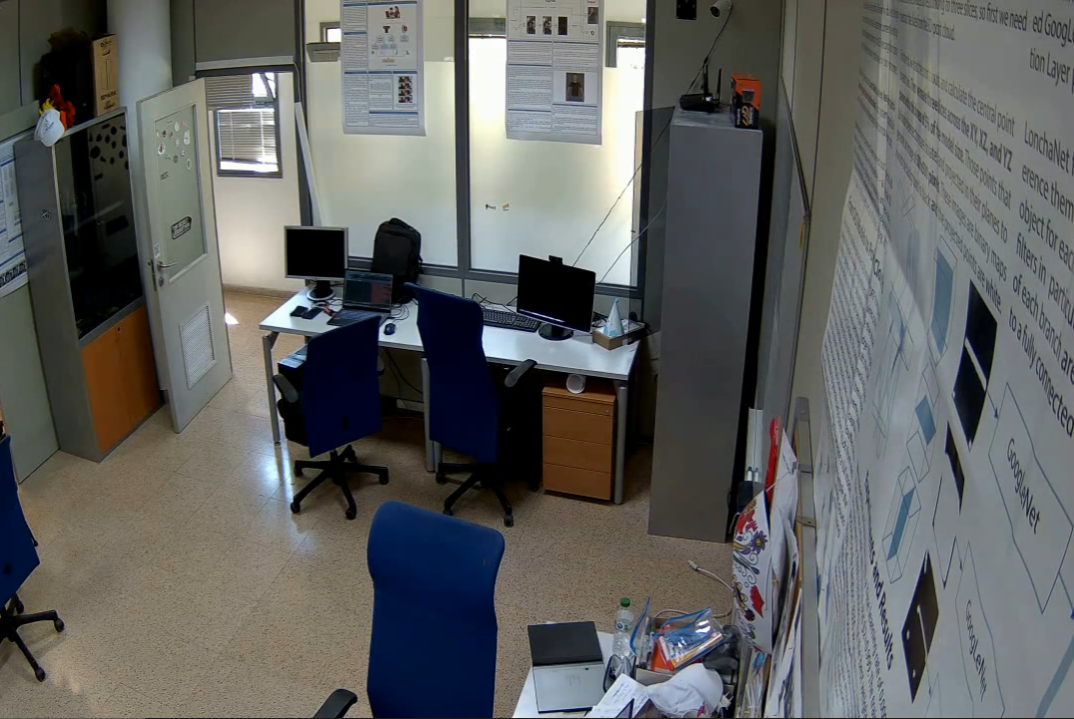}
    \includegraphics[width=0.24\linewidth, height=2.5cm]{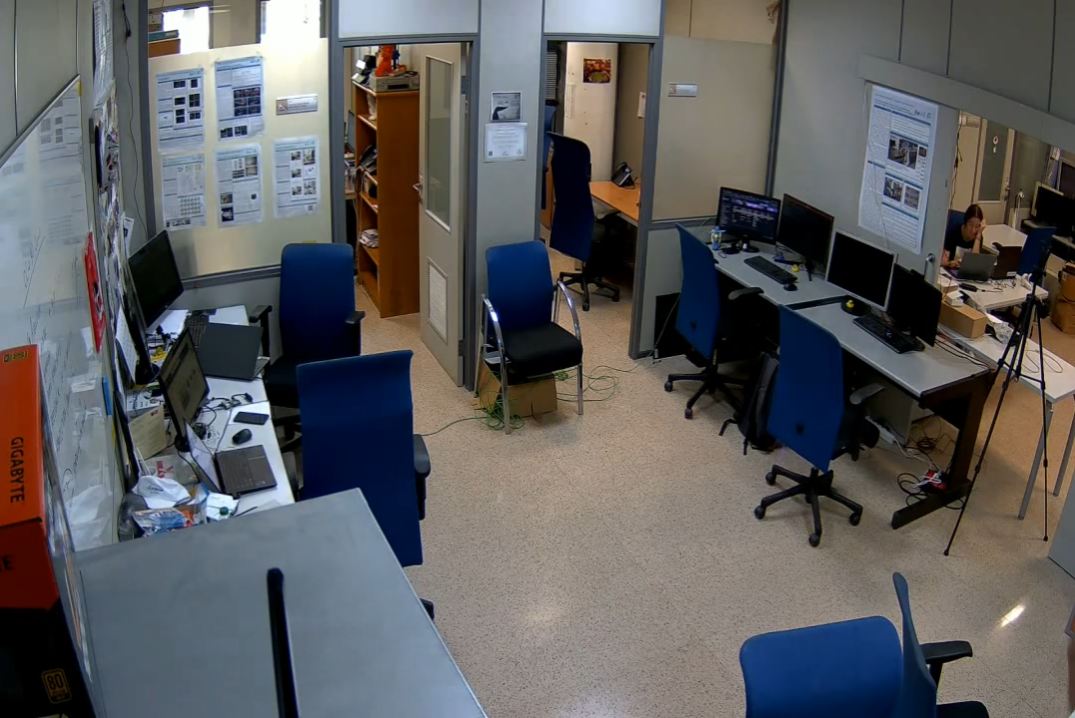}
    \includegraphics[width=0.24\linewidth, height=2.5cm]{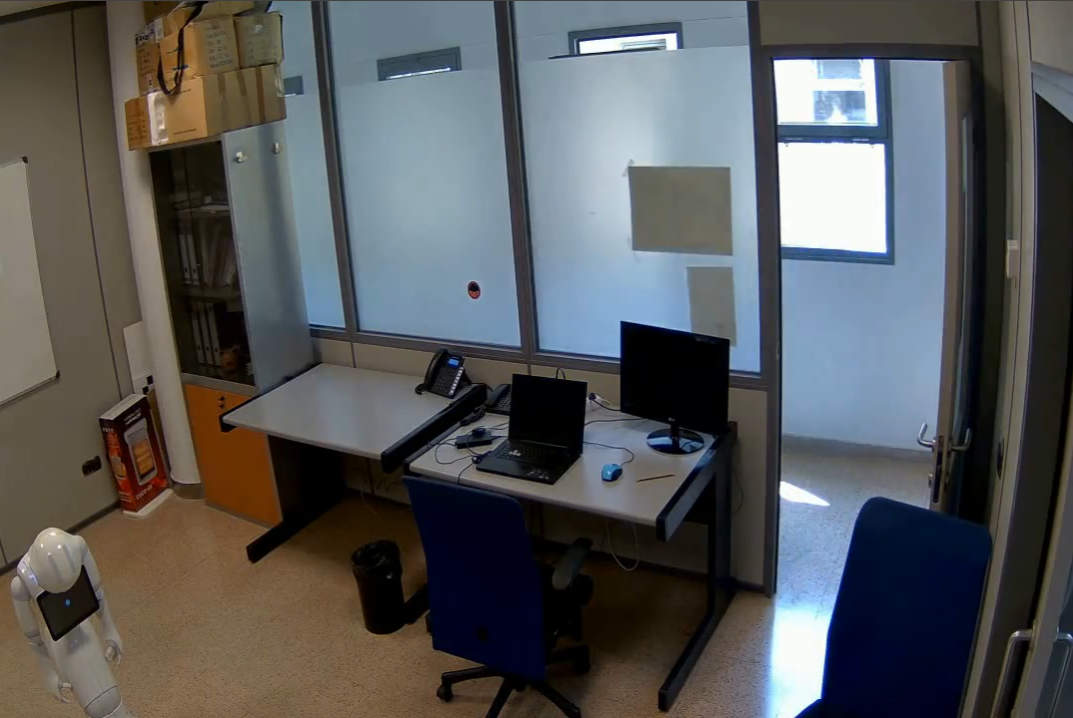}
    \includegraphics[width=0.24\linewidth, height=2.5cm]{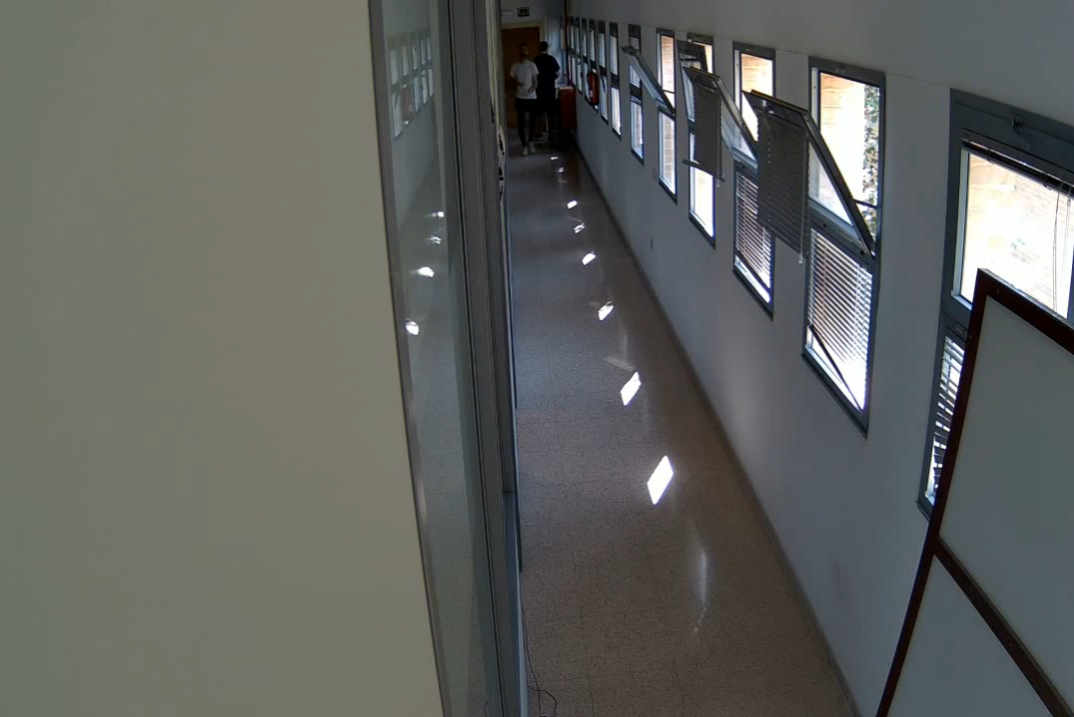}
    \includegraphics[width=0.24\linewidth, height=2.5cm]{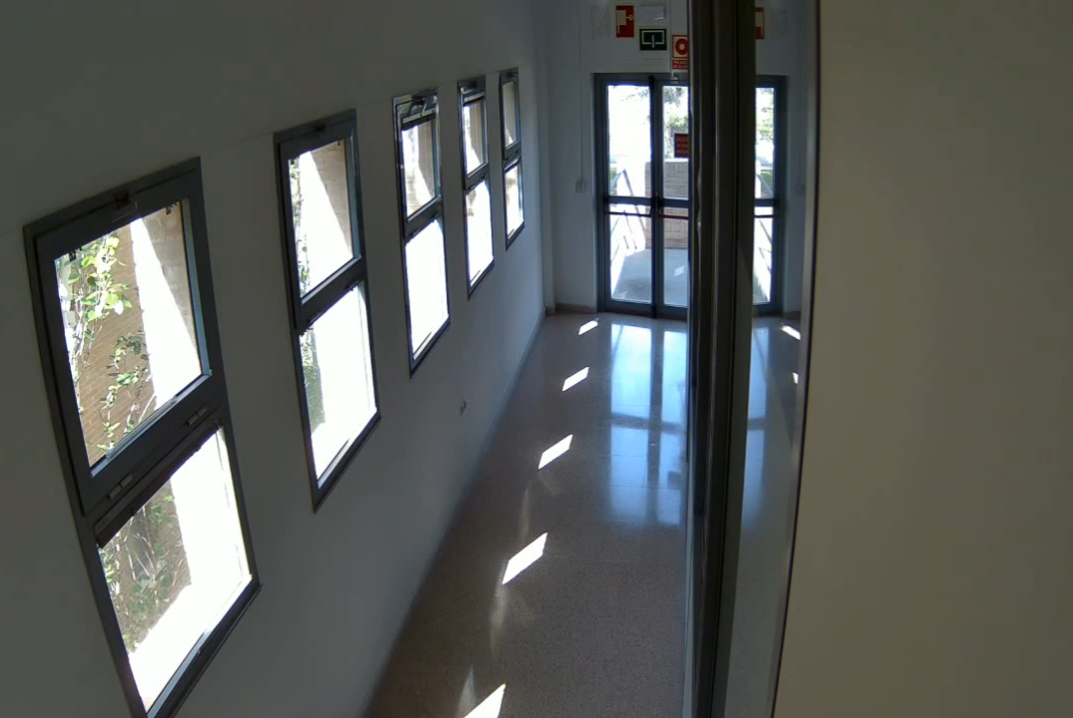}
    \includegraphics[width=0.24\linewidth, height=2.5cm]{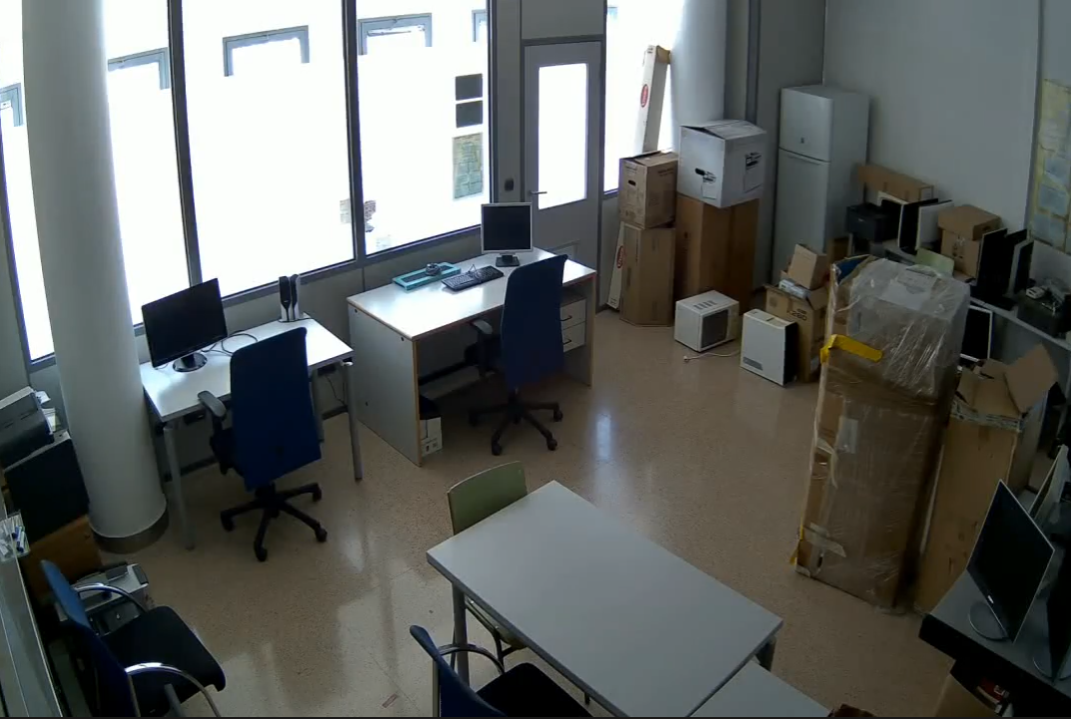}
    \caption{Sample frames of the point of view of each camera: C1-C4 (top row) and C5-C7 (bottom row).}
    \label{fig:camera_pov}
\end{figure}

The placements of the camera have been chosen, so the images depict most of the rooms and also the doors, which are of special interest for person Re-ID tasks.
Sample frames of the point of view of each camera are shown in Figure \ref{fig:camera_pov}. 

\subsection*{Hardware, Connection Architecture and Synchronization}
\label{subsec:synchro}

The hardware component of the system primarily consists of a network of cameras. The selected cameras are Reolink RLC-410W, which are equipped with a 1/2.7" CMOS image sensor and deliver 5.0 megapixel resolution. Each camera features a 4.0 mm lens with an aperture of F/2.0, providing an 80° horizontal and 58° vertical field of view. In addition, they include backlight compensation to enhance image quality. The video is compressed using the H.264 format. Despite the cameras' capability of recording at 2560×1920 pixels, videos were captured at 1080×720 pixels to prevent network congestion and maintain a consistent 30 fps across all cameras.

The camera network consists of seven cameras deployed in the environment. Although these cameras support WiFi connectivity, most of them were connected via Ethernet cables to ensure a faster and more stable connection. Six of the cameras were connected to a central router using Cat 6 Ethernet cables, while one camera (C7) was connected via WiFi. The selected router is the Asus RT-AC3100 Gigabit Router, which features 8 Ethernet ports and solid WiFi capabilities.

To complete the setup, a server computer was incorporated into the system. The chosen machine is powered by an Intel i9-13900K processor with 32 cores running at 5.5 GHz, paired with 32 GB of DDR4 RAM, and operates on Ubuntu 20.04. This server is responsible for running the interconnection framework, retrieving images from all cameras, synchronizing them, and storing the collected data.

To do all these tasks, we rely on ROS (Robotic Operating System) Noetic. ROS is a comprehensive collection of software libraries and tools designed to support the development of robotic applications. It offers everything from device drivers to advanced algorithms, along with a suite of powerful tools for developers. It also includes tools to connect to remote cameras, retrieve images from them, and allows the implementation of custom synchronization techniques.

To ensure synchronization among all the cameras and guarantee that they all capture the same moment in time, the process begins by waiting several seconds to establish stable connections with each camera. This delay ensures that all cameras are properly initialized and ready to operate without any interruptions. Additionally, the frame buffer in the camera software is disabled to prevent delays caused by accumulated frames, ensuring real-time visualization of the captured frames. Furthermore, a uniform capture frequency is configured across all cameras, allowing for consistent synchronization and ensuring that all cameras capture the exact same moment in time as closely as possible. This combination of techniques minimizes the risk of desynchronization and maximizes the accuracy of the system for time-sensitive applications.

Each camera is launched in parallel within a separate ROS node, utilizing OpenCV's functionality to connect to the video stream via the Real Time Streaming Protocol (RTSP). This setup ensures efficient management of multiple video streams, enabling the system to handle concurrent connections to each camera. By assigning each camera to an independent node, the architecture leverages the distributed nature of ROS, allowing for better scalability and resource management while maintaining the synchronization and real-time performance required for the application.

\subsection*{Procedure}
\label{subsec:procedure}

The procedure for recording a sequence is straightforward. The person in charge of the dataset started recording at a random moment of the day. The subjects present in the environment were instructed to move around without any specific purpose. No constraints were applied to their dynamics, interactions, or behaviors. This approach aimed to capture realistic scenarios, which involve multiple person occlusions, changes in trajectories, and even variations in clothing or appearance, posing a significant challenge to person Re-ID methods.

Nonetheless, the process of labeling the sequences involves both automatic labeling algorithms and manual annotation. 
First, individuals are extracted from each video using the YOLOv8 algorithm \cite{Jocher_Ultralytics_YOLO_2023}. YOLOv8 (You Only Look Once, Version 8) is a deep learning-based object detection architecture capable of predicting the location and class of specific objects within an image. It has different model sizes considering the depth of the architecture, the runtime and the accuracy it provides. The YOLOv8x (extra-large) version is used as it provides the best accuracy. In our case, we feed the architecture with all the frames in order to retrieve the regions occupied by persons. The actual output consists of the $X$ and $Y$ coordinates of the opposite corners of each person's bounding box. Despite YOLOv8 being a state-of-the-art object detection method, it is not flawless, so each automatically generated label was manually verified, being confirmed or corrected when necessary. Additionally, new labels were assigned to individuals who were missed by the automatic detection process.

It should be noted that individuals are labeled only if they can be sufficiently identified; otherwise, their labels are removed. In addition to the bounding box annotations, the dataset includes a unique ID for each person. These IDs were labeled following a semi-automatic approach based on Deep SORT \cite{deepsort}. Deep SORT is an advanced object tracking method that extends the SORT (Simple Online and Realtime Tracking) algorithm \cite{sort2016} by incorporating deep learning for more accurate tracking. It leverages appearance features extracted by a deep neural network to improve track association, allowing it to handle occlusions and re-identify objects more effectively across frames. The algorithm was applied to all videos in our dataset to retrieve tracklets, which were subsequently reviewed and corrected manually to ensure accuracy. Importantly, the assigned IDs remain consistent across all cameras and sequences.

\begin{figure}[!htb]
    \centering
    \includegraphics[height=6cm]{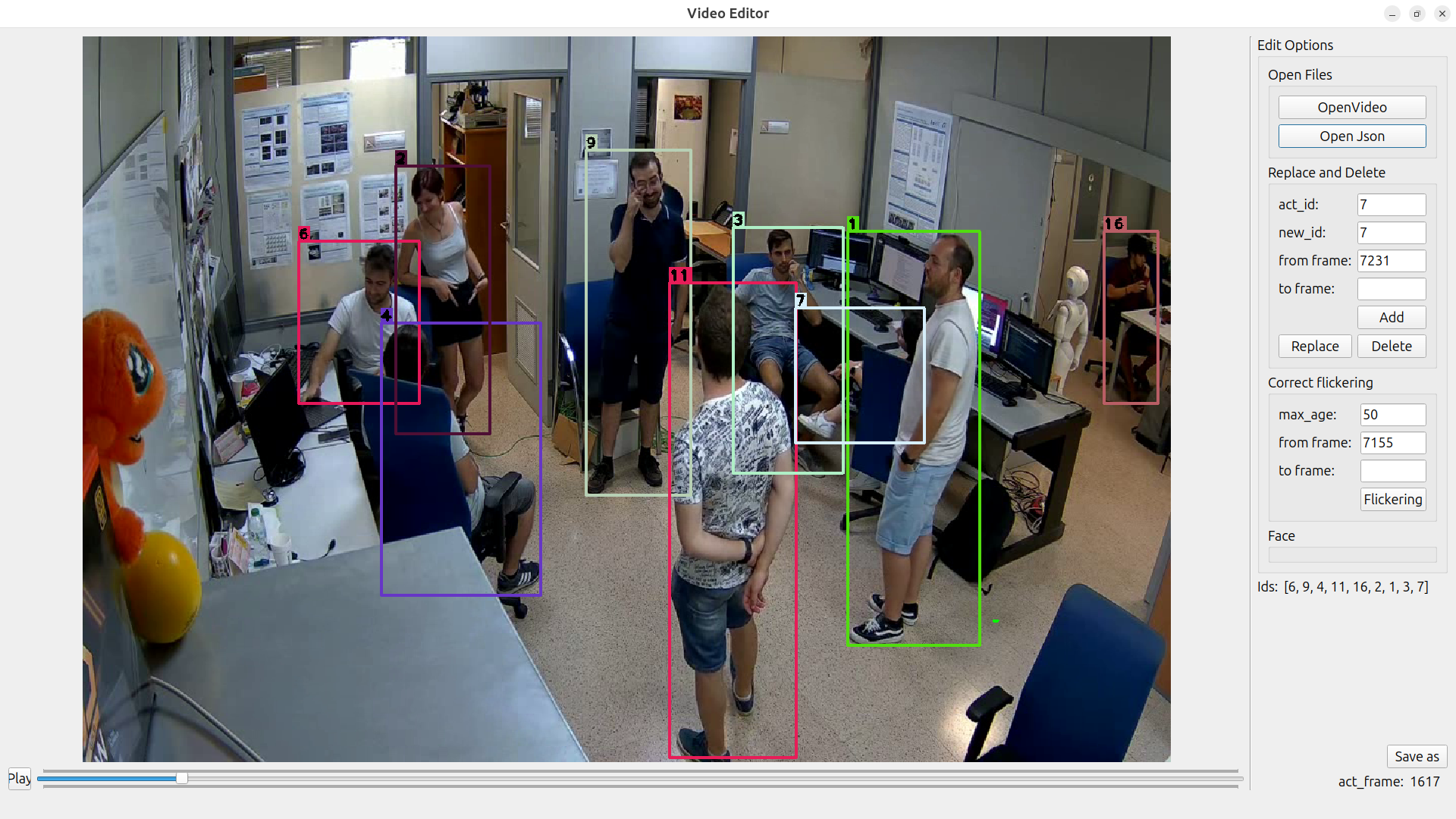}
    \caption{GUI designed to enhance annotations.}
    \label{fig:gui}
\end{figure}

To streamline and expedite the manual verification and correction of bounding boxes and ID annotations, we developed a graphical user interface (GUI). This tool significantly facilitates the labeling process (Figure \ref{fig:gui}). The GUI is publicly available at \url{https://github.com/bdager/preid-labeling-gui}.

\subsubsection*{Benchmark Data Preparation}

After preparing the dataset annotations, we used them to convert the data into a format required for the benchmarks proposed in Section Technical Validation. Different train and test splits were created for each benchmark scenario. Data were initially generated automatically following the procedures described below, after which a manual selection process was applied to determine the final train and test sets.


\paragraph{Tracking Benchmark Data Generation.}

For the \textbf{brief occlusions} tracking benchmark, we defined occlusions as ID trajectory interruptions lasting between one and five seconds. In this case, we included for each ID only the frames immediately before and after an occlusion began and ended, along with their corresponding bounding box coordinates.
For the \textbf{multiple-person occlusions} scenario, we extended the brief occlusion condition by requiring an intersection between the occluded ID and another ID before the occlusion occurred. We applied an Intersection over Union (IoU) threshold of 0.3 to ensure sufficient bounding box overlap before occlusion. 
From the generated data, we selected two sequences for training and eight for testing, prioritizing challenging scenarios in both tracking tasks.

\paragraph{Re-Identification Benchmark Data Generation.}

For person Re-ID benchmark, we focused on evaluating frames following long occlusions (lasting more than five seconds). This distinction ensures a clear differentiation from tracking scenarios that involve shorter occlusions.
In this case, we included in the train sets the first visible frame of each ID, followed by four consecutive frames sampled at an 18-frame gap (equivalent to 2.4 seconds of visibility). By contrast, the test sets comprise the reappearance frames and the four subsequent frames for each reappearance, using the same sampling strategy as the train sets. However, a minimum of 20 frames per ID is required in the test sets. If fewer than 20 frames are available, additional frames are randomly sampled to meet the requirement.

In addition, for IDs present in the dataset, we assigned negative values in the test sets when they meet the long-term occlusion conditions but are not included in the train sets. This labeling explicitly defines them as unknown IDs, what we consider as distractors, providing an additional challenge to the Re-ID system.

\begin{figure}[!htb]
    \centering
    \begin{subfigure}[b]{0.2\textwidth}
        \centering
        \includegraphics[height=5.2cm]{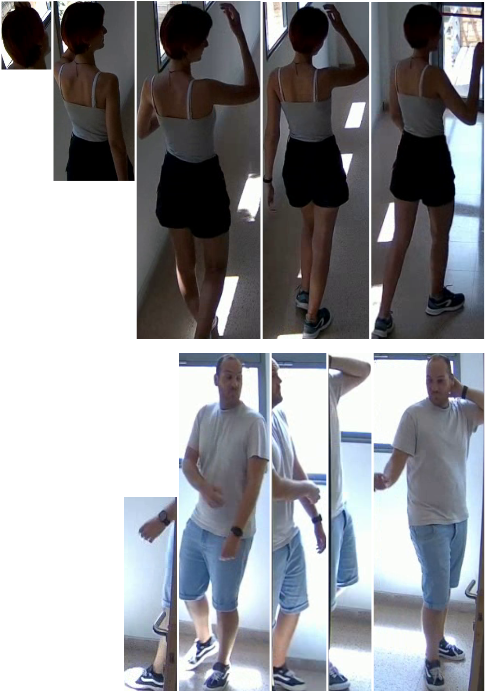}
        \caption{Train data samples}
        \label{fig:train_samples}
    \end{subfigure}
    \hspace{0.04\textwidth} 
    \begin{subfigure}[b]{0.45\textwidth}
        \centering
        \includegraphics[height=5.2cm]{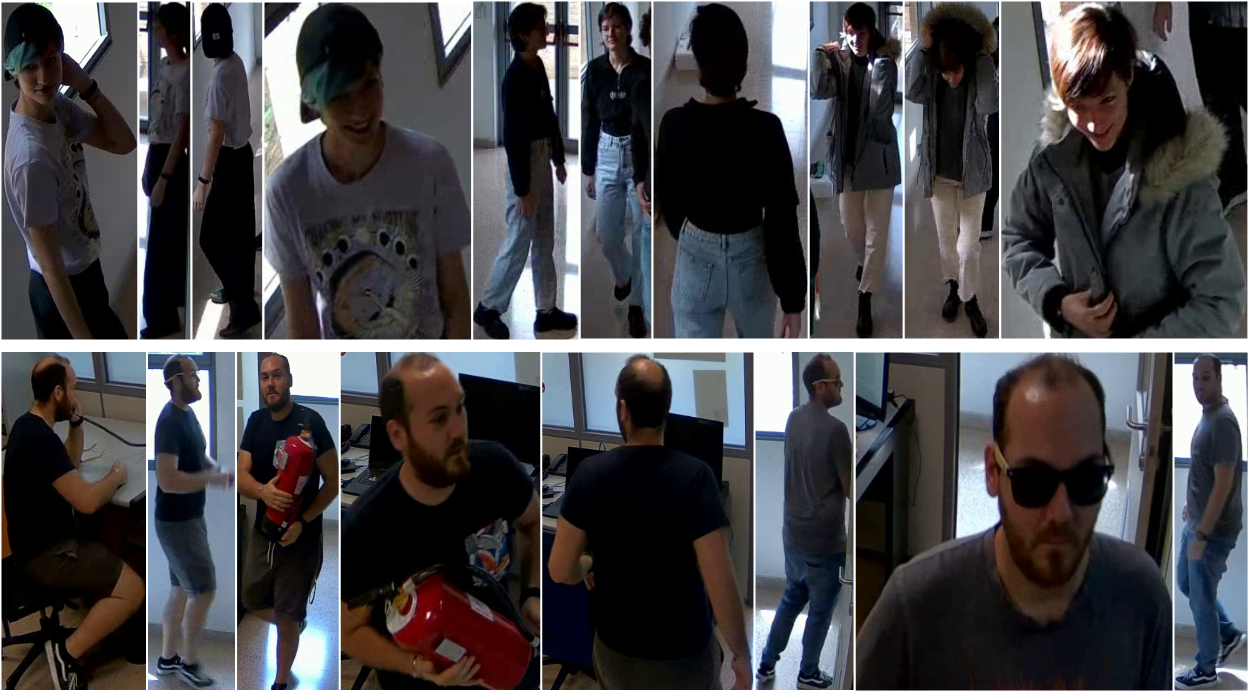}
        \caption{Test data samples}
        \label{fig:test_samples}
    \end{subfigure}
    \caption{Samples of train and test data for two IDs in the long-term Re-ID benchmark. The first row represents ID "2" from the CHIRLA dataset, captured by camera 5 at different time intervals. The second row represents ID "1", captured by camera 4 over time. These samples demonstrate the variations in clothing and appearance of individuals across different time periods.}
    \label{fig:train_test}
\end{figure}

In Section Technical Validation, the Re-ID benchmark is divided into four scenarios. For each, we selected subsets of train and test data, with the goal that the final scenario evaluation is computed as the average of individual subset results. All evaluation data was carefully analyzed to align with benchmark definitions, emphasizing challenging cases. Figure \ref{fig:train_test} illustrates sample data from one Re-ID benchmark scenario.

\begin{itemize}
    \item \textbf{Re-identification after reappearance on the scene.}  
    We manually selected 10 train and test splits. The train set consists of annotations from 10 videos, each subset containing one video per sequence. However, only nine distinct sequences met the long occlusion condition, requiring the repetition of one sequence with a different sampled video. If uninterrupted frames were unavailable within the analyzed duration (2.4 seconds), only frames within the specified range were included. Each test set included all frames where the individual reappeared after a long occlusion.

    \item \textbf{Long-term re-identification.}  
    We selected seven train and test subsets. The train set included annotations from one video per camera, totaling seven videos, with one per subset. The corresponding test set comprised all remaining videos from the same camera analyzed in the train set across different sequences.

    \item \textbf{Multi-camera re-identification.}  
    Data was distributed across 10 train and test subsets. The train set consisted of annotations from one video per sequence, totaling 10 videos, with one per subset. The corresponding test subset included all remaining videos from the same sequence.

    \item \textbf{Multi-camera long-term re-identification.}  
    Similar to the previous scenario, we selected 10 subsets for training and testing. The train set consisted of annotations from one video per sequence, totaling 10 videos. Each test subset contained two videos from one or two sequences different from those in the train subset.
\end{itemize}


\subsection*{Evaluation Metrics}
\label{subsec:evaluation}

This dataset can be used for evaluation from two different but complementary points of view: pure Re-ID of persons and tracking of persons, allowing both, algorithms that perform only one of the two tasks and those that incorporate both facets to be evaluated.

\subsubsection*{Person Tracking Metrics}
\label{subsec:people-tracking-metrics}

In terms of person tracking metrics, which assess the ability of algorithms for short-term recognition, i.e., to detect a person, identify them and follow them while they remain in the scene until they leave it. In this respect, we will rely on the CLEAR MOT Metrics \cite{bernardin2008evaluating, mot16, deepsort} to define the following metrics to evaluate the tracking of persons on individual cameras since the areas covered by each camera do not overlap each other, or the percentage of overlap is low.

To understand the following metrics, we must define some concepts:
\begin{itemize}
\item {\textbf{Fragmentation ($FM$)}.  The number of interruptions in a ground truth trajectory. 
A fragmentation is recorded each time a trajectory's status changes from "tracked" to "untracked" and tracking of the same trajectory is resumed at a later point.} 
\item {\textbf{Identity Switch ($IDSW$)}. The number of times the reported ID associated with a ground truth trajectory changes throughout the tracking process. This occurs when the tracking system incorrectly reassigns the ID of a tracked person to a different ID. } 
\item {\textbf{Mostly Tracked ($MT$)}. The percentage of ground truth trajectories that correctly assigned the same ID label for at least 80\% of their lifespan. } 
\item {\textbf{Mostly Lost ($ML$)}. The percentage of ground-truth trajectories that are successfully tracked for no more than 20\% of their lifespan. } 

\item {\textbf{False Positive ($fp_t$)}. The number of targets the tracker detects in frame $t$ where there is none in the ground truth.}
\item {\textbf{False Negative ($fn_t$)}. The number of true targets missed by the tracker in frame $t$.}
\end{itemize}

The most used metric to evaluate multi-object tracking performance for a single-camera is Multiple Object Tracking Accuracy (MOTA), which is defined as shown in Equation \ref{eq:mota}.

\begin{equation}
\label{eq:mota}
    MOTA = 1 - \frac{FN + FP + IDSW}{GT},
\end{equation}

where:
\begin{conditions}
FN & total number of false negatives over all frames. \\
FP & total number of false positives over all frames. \\
IDSW & total number of ID switches over all frames. \\   
GT & total number of ground truth detections over all frames.
\end{conditions}

MOTA can be seen as derived from three error ratios:
\begin{itemize}
    \item The ratio of false negatives in the sequence, computed over the total number of objects present in all frames, shown in Equation \ref{eq:misses}.
    \item The ratio of false positives, shown in Equation \ref{eq:false-positives}.
    \item The ratio of ID switches, shown in Equation \ref{eq:mismatches}.
\end{itemize}

\begin{equation}
\label{eq:misses}
    {FN}_{ratio} = \frac{FN}{GT},
\end{equation}

\begin{equation}
\label{eq:false-positives}
    {FP}_{ratio} = \frac{FP}{GT},
\end{equation}

\begin{equation}
\label{eq:mismatches}
    {IDSW}_{ratio} = \frac{IDSW}{GT}.
\end{equation}

MOTA takes into account all object configuration errors made by the tracker across all frames: false positives, false negatives and ID switches. %
It provides a very intuitive measure of the tracker's performance in detecting objects and maintaining their trajectories, regardless of the accuracy with which object locations are estimated. 

A complementary metric is IDF1, which evaluates the consistency of object IDs over time  \cite{ristani2016performance}. Specifically, IDF1 measures the F1-score of correctly matched object trajectories, using intersection over union (IoU) and a threshold, considering both the precision and recall of the identifications at the frame level. IDP (ID Precision) measures the precision of detected trajectories that are correctly matched with the ground truth trajectories while IDR (ID Recall) measures the recall of ground truth trajectories that are correctly matched with the detected trajectories. The final IDF1-Score is the harmonic mean of IDP and IDR, calculated as shown in Equation \ref{eq:idf1}.

\begin{equation}
\label{eq:idf1}
IDF1 = 2 * \frac{IDP * IDR}{IDP+IDR}.
\end{equation}

IDF1 is particularly important because it evaluates how well an algorithm maintains the consistency of object IDs across frames. Unlike MOTA, IDF1 places additional emphasis on the quality of ID matching.

\subsubsection*{Re-identification Metrics}
\label{subsec:reid-metrics}

Tracking metrics evaluate a method's ability to maintain the same ID throughout a sequence once it has been assigned. However, these metrics do not account for the system's ability to re-identify individuals across different sequences, particularly when a person leaves the scene and a significant amount of time has passed before their reappearance. In this section, short-term tracking capability is set aside, and the emphasis is placed on the system's ability to recognize IDs across different cameras, viewpoints, and noticeable changes in appearance.

The Cumulative Matching Characteristic (CMC) is a widely used evaluation metric in person Re-ID tasks for assessing the performance of recognition algorithms \cite{gray2007evaluating}. The CMC curve represents the probability that a correct match appears within the top $k$ ranks of a ranked list of matches. This metric provides a function of different $k$-rank accuracies, as defined in Equation \ref{eq:cmc}.

\begin{equation}
\label{eq:cmc}
     Acc_k= \left\{ \begin{array}{ll} 1, & \text{if top-$k$ ranked gallery samples contain the query ID} \\ \\ 0, & \text{otherwise} \end{array} \right.
\end{equation}

In a typical person Re-ID scenario, the system is given a query image of a person, and it has to find the correct match from a gallery of images. The system ranks all images in the gallery based on similarity to the query image. The CMC curve plots the probability of the correct match being within the top $k$ ranked positions across all queries. This metric is suitable for scenarios where the task is to rank potential matches rather than to assign a binary label to them. This ranking approach is crucial because, in large datasets, the probability of a random match being correct is very low, making it an ineffective metric for recognition. The CMC curve addresses this by focusing on ranking performance rather than binary classification performance. One of the most important associated metrics is the rank-$1$ accuracy, which means the system can correctly identify an individual from the probe camera in the gallery camera as the top match. 

CMC evaluation measurement is valid only if there is only one ground truth match for a given query \cite{zheng2015scalable}. However, if multiple ground truths exist, the CMC curve can be biased because recall is not considered. Therefore, an alternative metric to evaluate the overall performance of the Re-ID system is the Mean Average Precision (mAP). For each query, the area under the Precision-Recall curve, referred to as average precision (AP), is computed. Precision can be defined as the ratio between correct ID correspondences (True Positives) and the number of retrieved correspondences, as shown in Equation \ref{eq:precision}.

\begin{equation}
\label{eq:precision}
    precision = \frac{TP}{TP+FP}.
\end{equation}

By default, precision takes all the retrieved correspondences into account, but it can also be evaluated at a given number of retrieved images, commonly known as cut-off rank, where the model is only assessed by considering its top-most queries. The measure is called precision at $k$ or P$@k$. Using the P$@k$ we can calculate the average precision (AP$@n$) for a number $n$ of matches, as shown in Equation \ref{eq:apn}. The final AP is the value of AP$@n$ when all the ground truth positives have been obtained.

\begin{equation}
\label{eq:apn}
    AP@n = \frac{1}{GTP} \sum_{k=1}^{n}P@k * rel@k.
\end{equation}

\noindent where:
\begin{conditions}
GTP & total number of ground truth positives (correct samples of the query class), \\
$n$ & total number of candidate matches ($n$ elements most similar to the query), \\
rel@k  & relevance function that equals 1 if the ID at rank $k$ is correct. \\
\end{conditions}

Subsequently, the mean of all AP values across all queries, known as mAP, is calculated, as shown in Equation \ref{eq:map}. This metric takes into account both the precision and recall of an algorithm, offering a more thorough evaluation.

\begin{equation}
\label{eq:map}
    mAP = \frac{1}{N}\sum_{i=1}^{N} AP_i.
\end{equation}

\noindent where:
\begin{conditions}
N & total number of queries, \\
i & index for the i-th query, 
\end{conditions}

\section*{Data Records}
\label{sec:data-records}




The dataset is composed of a collection of videos, as provided by the set of cameras in the environment, the corresponding Re-ID labels, and train and test data to use in the benchmarks proposed in Section Technical Validation. The Re-ID labels comprise the bounding box of the individuals present in each frame of the video and the corresponding ID. In the following subsections, we will delve into the components of the dataset and how it is organized.

\subsection*{Dataset Components}
\label{sec:dataset_components}

The dataset consists of 10 sequences, each containing seven videos recorded by the seven cameras deployed in the environment. The videos were collected over a period of seven months, allowing the dataset to capture variations in both the environment and the appearance of individuals. The average duration of a single video is 284 seconds, with a range of 113 to 567 seconds. In total, the dataset comprises more than 5.52 hours of video and 596,345 frames, with a total of 963,554 annotated bounding boxes. The details of each sequence are presented in Table \ref{tab:seq}. The videos are stored in AVI format, encoded using DivX MPEG-4 at 30 fps with a bitrate of 4048 Kbps.

\begin{table}[!htb]
    \centering
    {
    \begin{tabular}{|l|p{6cm}|l|l|l|}
    \hline
    Sequence ID & Subject IDs & Duration (s) & Avg. \# Frames & \# Bboxes \\ \hline
    0           &  1, 3, 5, 6, 9, 10, 18, 19                   & 298   &  8,913   &  77,510  \\ \hline
    1           &  1, 3, 5, 6, 8, 9, 10, 18, 19                & 131   &  3,894   &  50,374  \\ \hline
    2           &  1, 3, 5, 6, 8, 9, 10, 18, 19                & 113   &  3,351   &  43,058  \\ \hline
    4           &  1, 2, 3, 4, 5, 6, 7, 8                      & 286   &  8,565   &  82,430  \\ \hline
    6           &  1, 2, 3, 4, 5, 9, 24                        & 205   &  6,237   &  55,075  \\ \hline
    7           &  1, 2, 3, 4, 5, 9                            & 250   &  7,504   &  66,451  \\ \hline
    20          &  1, 2, 3, 4, 6, 7, 9, 10, 11, 12, 13, 15, 16 & 463   &  13,903  &  203,143 \\ \hline
    24          &  2, 5, 6, 7, 10, 11, 14, 25, 26              & 567   &  16,988  &  214,792 \\ \hline
    25          &  2, 6, 7, 9, 10, 12, 14                      & 270   &  8,084   &  69,432  \\ \hline
    26          &  1, 2, 3, 5, 6, 7, 9, 10, 11, 14             & 258   &  7,749   &  101,289 \\ \hline
    \end{tabular}
    }
    \caption{Data from the sequences of CHIRLA Dataset.}
    \label{tab:seq}
\end{table}

The dataset includes 22 different individuals, 17 males and five females, ranging in age from 20 to 60 years. As expected, each ID has distinct physical features, and their appearance may vary across different sequences. However, it is worth noting that not all individuals appear in every sequence, nor are they necessarily visible in all cameras.

In addition to the video data, we provide Re-ID labels.  As mentioned earlier, each video is accompanied by a label file that contains, for each frame, the bounding boxes of all detected individuals along with their corresponding IDs. The labels are stored in JSON format, making them easy to parse.
Each JSON file consists of a dictionary where the keys represent the frame numbers of the video, and the corresponding values are lists. Each element in these lists is a dictionary containing two keys: 

\begin{itemize}
\item \texttt{"id"}: Unique ID assigned to each person.
\item \texttt{"BboxP"}: A list of the bounding box coordinates \([X_1, Y_1, X_2, Y_2]\) that define the region occupied by the person in the analyzed frame.
\end{itemize}

\begin{figure}[!htb]
    \centering
    \includegraphics[width=0.32\linewidth, height=3.5cm]{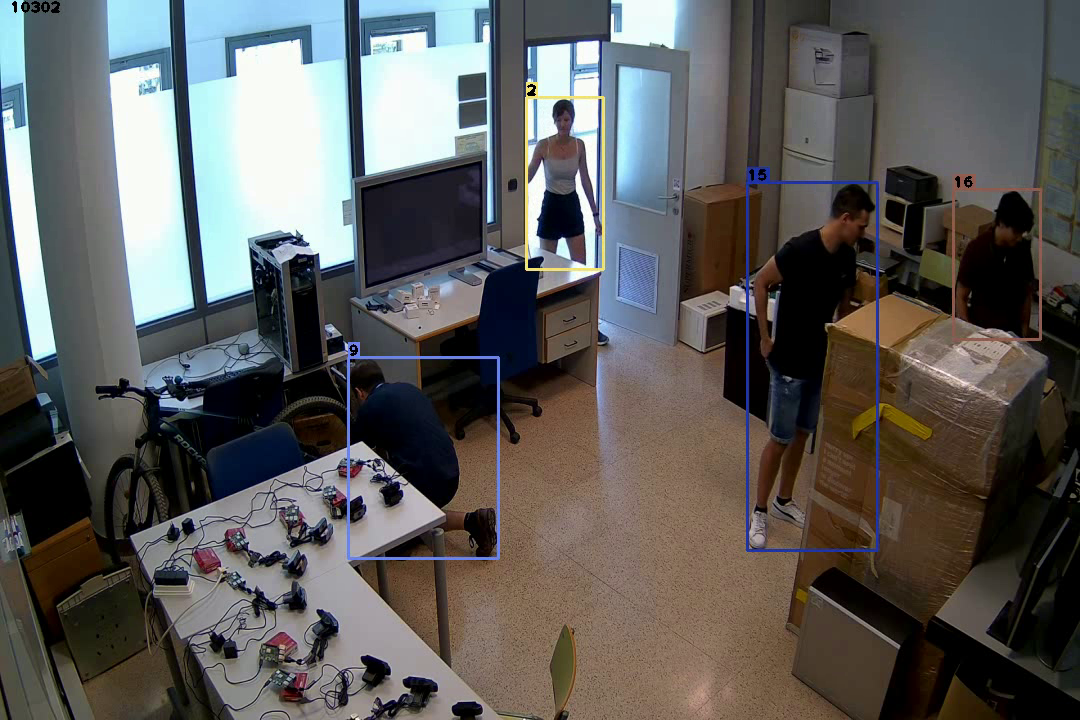}
    \includegraphics[width=0.32\linewidth, height=3.5cm]{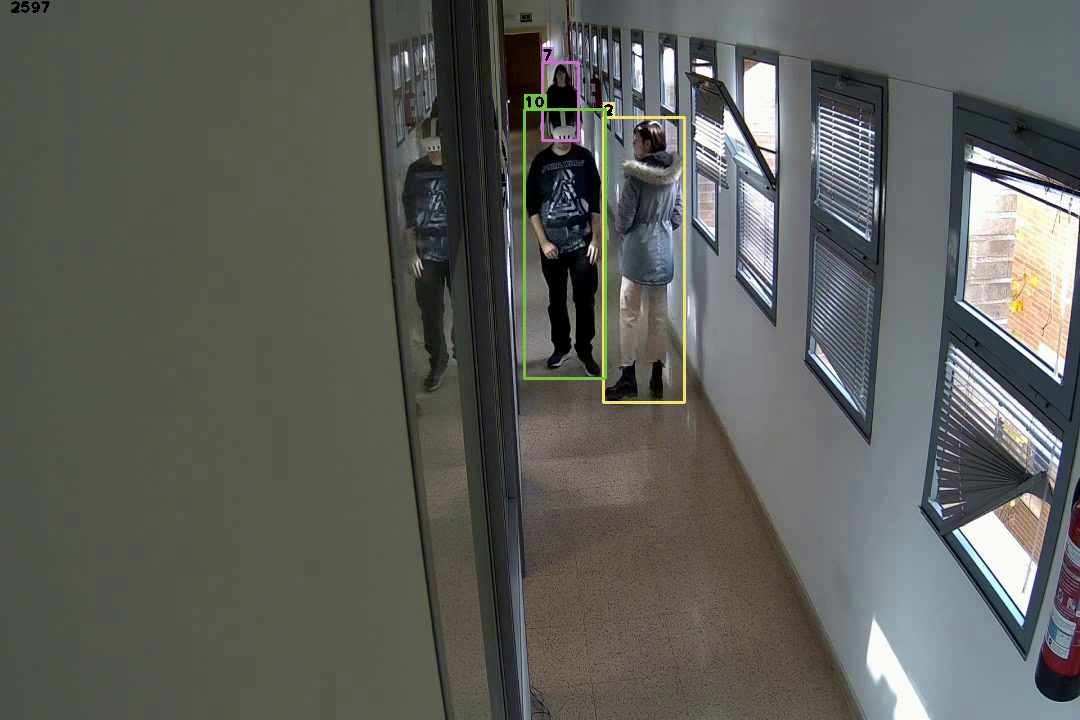}
    \includegraphics[width=0.32\linewidth, height=3.5cm]{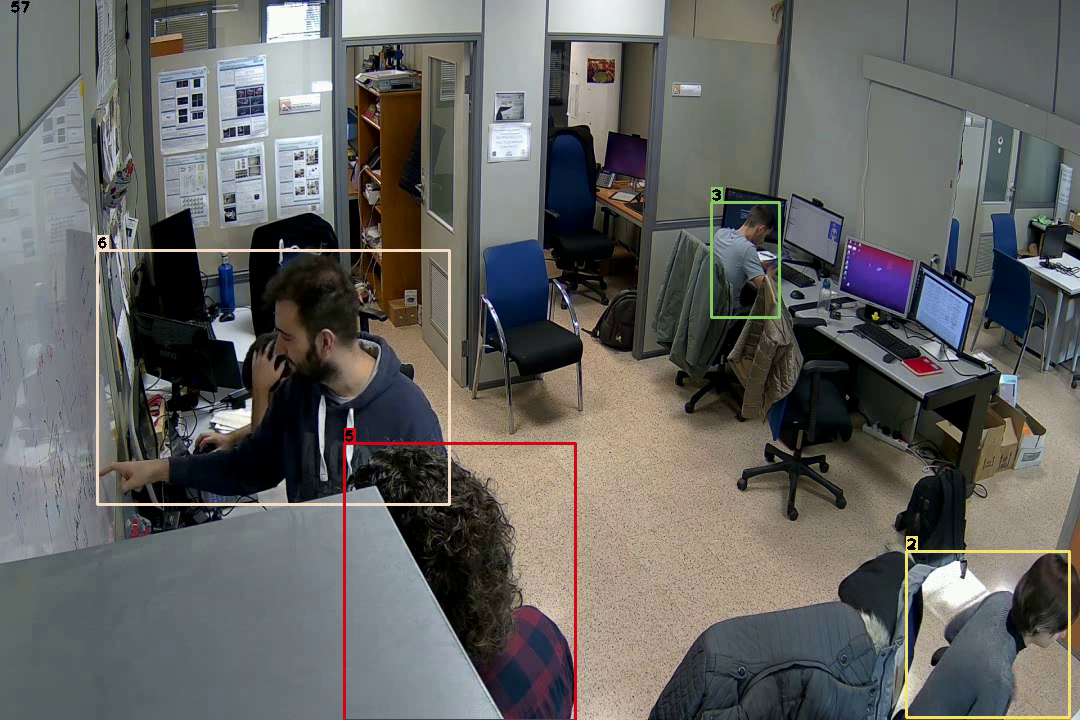}
    \caption{Random frames with the labels superimposed.}
    \label{fig:labels_sample}
\end{figure}

Frame and ID values are 1-based, whereas bounding box coordinates are 0-based. Specifically, the top-left corner of the bounding box corresponds to $(0,0)$. Figure \ref{fig:labels_sample} presents examples of labeled individuals in randomly selected frames from the dataset.


\subsubsection*{Train and Test Data}

Besides the video and Re-ID data, the dataset also includes train and test data specifically selected to use in the benchmarks proposed in Section Technical Validation. 
For the tracking benchmark, the dataset provides data for two tasks: \textbf{Tracking after brief occlusions} and \textbf{Tracking in scenarios with multiple people and occlusions}. For each scenario, the train set consists of two sequences, while the test set includes eight sequences. 
In both cases, the data is structured in a hierarchical JSON format. Each JSON file consists of a dictionary where the keys represent individual IDs in the video, and the values are lists of annotations. Each annotation is represented as a dictionary containing four keys:
\begin{itemize}
    \item \texttt{"frame\_init"}: The frame number immediately before the occlusion starts.
    \item \texttt{"frame\_end"}: The frame number immediately after the occlusion ends.
    \item \texttt{"bbox\_init"}: A list of coordinates \([X1, Y1, X2, Y2]\) defining the bounding box of the person in the frame before the occlusion starts.
    \item \texttt{"bbox\_end"}: A list of coordinates \([X1, Y1, X2, Y2]\) defining the bounding box of the person in the frame after the occlusion ends.
\end{itemize}

For person Re-ID benchmark, same as in tracking benchmark, all data is also organized in a hierarchical JSON structure. Each JSON file consists of a dictionary where the keys represent individual IDs within the video, and the values are lists of annotations. Each annotation is represented as a dictionary with the following two keys:
\begin{itemize}
    \item \texttt{"frame"}: The frame number where the person is present in the video.
    \item \texttt{"BboxP"}: A list of coordinates \([X1, Y1, X2, Y2]\) representing the bounding box of the person in that frame, where \([X1, Y1]\) denotes the top-left corner, and \([X2, Y2]\) denotes the bottom-right corner.
\end{itemize}

The dataset provides train and test splits for four Re-ID benchmark scenarios. The \textbf{Re-identification after reappearance on the scene} scenario includes 10 train and test splits, each containing one video per sequence, selected from nine distinct sequences. The \textbf{Long-term re-identification} scenario consists of seven subsets, with the train set containing one video per camera and the test set comprising all remaining videos from the same camera. The \textbf{Multi-camera re-identification} scenario is structured into 10 train and test subsets, with each train set including one video per sequence and the corresponding test set containing all remaining videos within the same sequence. Lastly, the \textbf{Multi-camera long-term re-identification} scenario also includes 10 train and test subsets, where each test subset contains two videos from one or two sequences distinct from those in the train set.

Finally, the dataset is publicly available in the Science Data Bank data repository and can be accessed following the link \url{https://doi.org/10.57760/sciencedb.20543}.

\subsection*{Organization}

The dataset is structured into three primary directories: \texttt{annotations}, \texttt{videos}, and \texttt{benchmark}. The first two directories contain 10 subdirectories corresponding to the individual sequences listed in Table \ref{tab:seq}. Each sequence is labeled as \texttt{seq\_XXX}, where \texttt{XXX} represents a three-digit, zero-padded sequence number (e.g., \texttt{seq\_026}). Within each sequence subdirectory, data is further organized by camera views, with seven files per sequence—one per camera. Each file is named based on the camera number, date, and time of recording.

The \texttt{videos} directory contains the video files specific to each camera view and sequence. Each sequence subdirectory includes seven video files, one for each camera view, stored in AVI format (e.g., \texttt{camera\_1\_2023-12-05-12:12:13.avi}). 
The \texttt{annotations} directory contains the corresponding annotation files in JSON format. Each sequence includes seven JSON files, one per camera view. Each file provides frame-wise annotations for the corresponding camera view within the sequence, including bounding box coordinates and object IDs.


The \texttt{benchmark} directory contains data for benchmarks proposed in Section Technical Validation. This directory is divided into two main subdirectories: \texttt{tracking} and \texttt{reid}, each containing subdirectories for different benchmark scenarios, further split into train and test sets.


The \texttt{tracking} subdirectory consists of two benchmark scenarios: \texttt{brief\_occlusions} and \texttt{multiple\_people\_} \texttt{occlusions}. Each scenario is divided into \texttt{train} and \texttt{test} subdirectories, organized by sequence name (e.g., \texttt{seq\_004}).  
Within each sequence subdirectory, data is further organized by camera views. Each sequence contains seven JSON files (one per camera), labeled with the camera name and metadata (e.g., \texttt{camera\_7\_2023-06-08-12:29:08.json}). Alongside these JSON files, an \texttt{imgs} directory is provided, containing seven subdirectories—one per camera view—named according to the corresponding JSON files.  
Inside each camera view subdirectory, data is structured into subfolders named after the object IDs in the corresponding video. Each ID subfolder contains cropped bounding box images of that ID, selected based on the tracking scenario. These images are stored in PNG format and named as \texttt{frame\_n.png}, where \texttt{n} represents the frame number in the original video.


The \texttt{reid} subdirectory contains four re-identification scenarios: \texttt{reappearance}, \texttt{long\_term}, \texttt{multi\_camera}, and \texttt{multi\_camera\_long\_term}. Each scenario is split into \texttt{train} and \texttt{test} subdirectories, further divided into subsets named \texttt{train\_X} or \texttt{test\_X}, where \texttt{X} denotes the subset index (e.g., \texttt{train\_0}, \texttt{test\_1}). The \texttt{long\_term} scenario consists of seven subsets in both train and test sets, whereas the other three scenarios (\texttt{reappearance}, \texttt{multi\_camera}, and \texttt{multi\_camera\_long\_term}) contain 10 subsets each.
The directory structure in the re-identification benchmark follows the same format as in the tracking benchmark, including sequence-based organization, JSON annotation files, and \texttt{imgs} directories. 

\section*{Technical Validation}
\label{sec:tech-val}

In this section, we present well-defined benchmarks, to rigorously tested our dataset to ensure it meets the necessary requirements for challenging and realistic scenarios in person tracking and Re-ID systems. 
We present a series of experiments designed to evaluate the robustness of the algorithms in diverse settings. 

\subsection*{Benchmark}
\label{sec:benchmark}

Here, we describe the proposed tests designed to evaluate both the tracking and Re-ID systems, as well as those that integrate both functionalities. Each system will be individually assessed against the two benchmarks defined in the following subsections.

\subsubsection*{Benchmark for Person Tracking}
\label{sec:benchmark-tracking}

These tests assess the ability of a system to detect a person in the scene, create an ID for them, and track them on a single camera until the person disappears from the scene for a few seconds. 
For this purpose, the video sequences are divided into a train set and a test set. The train set is used to optimize certain parameters of the tracking algorithm that make it applicable to this setting, and adapt them to the resolution and the distance at which persons are seen in the scene. On the other hand, the test set will be used to evaluate the performance of the tracking systems, making use of the MOTA, IDF1 and their auxiliary metrics, as well as the generalization capacity of these systems after having optimized their parameters in videos from train sequences only.

With the proposed train and test sets, we check the capabilities of the tracking system in the following scenarios:

\begin{itemize}
    \item \textbf{Tracking after brief occlusions}. The person is totally hidden by an object or another person for a short period of time and becomes visible again without having left the scene. The objective of this scenario is to assess the ability of the system to maintain the ID and tracking of a person after a brief occlusion. 
    \item \textbf{Tracking in scenarios with multiple people and occlusions}. The scene includes several people, some of whom cross each other or are hidden, creating frequent occlusions. The objective of this scenario is to evaluate the system's ability to track multiple people simultaneously, even in situations where occlusions arise from their interactions.

In the analyzed scenarios, occlusions are treated as total occlusions, meaning no detections of the analyzed person are available in the occluded frames.
\end{itemize}

\subsubsection*{Benchmark for Re-ID}
\label{sec:benchmark-reidentification}

These benchmarks evaluate the ability of a system to re-identify a person in different situations that affect the continuity of their identification on one or more cameras. The objective is to determine the system's ability to recognize a person who has left the field of view or whose appearance has changed, based on comparison with previous images or sequences.

For this purpose, the video sequences are divided into train and test sets. The train sets are used to optimize the parameters of the Re-ID algorithm, adapting it to the visual characteristics of the recorded persons, so that it learns to recognize their ID, especially focusing on the characteristics that allow them to be identified in the long term. 

On the other hand, the test sets are designed to evaluate the performance of the system in different Re-ID scenarios, using metrics such as mAP, CMC and other auxiliary metrics that allow measuring its generalization capacity after the optimization of parameters in sequences other than train ones.

With the proposed train and test sets, we check the capabilities of the Re-ID system in the following scenarios:

\begin{itemize}

    \item \textbf{Re-identification after reappearance on the scene}. The person leaves the field of view of a camera and returns to the scene after a time interval. This reappearance can also include a change in the person appearance (e.g., takes off or puts on a jacket, wears a hat or changes an accessory). 
    The objective of this scenario is to assess the ability of the system to re-identify the person correctly after a break in visibility.
    \item \textbf{Long-term re-identification}. The person is seen in different recordings captured on different days, implying possible changes in appearance and environment. The objective of this scenario is to evaluate the system's ability to recognize a person when recordings come from different days, with potential variations in lighting, appearance and other contextual factors.
    \item \textbf{Multi-camera re-identification}. The person is captured by several cameras at different locations, with little or no overlap between the fields of view of the cameras. 
    The objective of this scenario is to assess the system's ability to re-identify the person when viewed from different angles and perspectives.
    \item \textbf{Multi-camera Long-term re-identification.} The person is captured by several cameras at different locations and days. The objective of this scenario is to evaluate the system's robustness in re-identifying the person from various angles and perspectives, ensuring reliable performance under diverse temporal and contextual conditions.
\end{itemize}

\subsubsection*{ {Benchmark Evaluation}}

{ 

In Table \ref{tab:track-eval} we report the results of different tracking methods evaluated on the CHIRLA benchmark for person tracking under the two proposed scenarios. We first present the results on the full test set (8 sequences with a total of 56 videos) using the metrics described in Section Person Tracking Methods, which reflect the general performance of person tracking across the dataset. We then provide results for the two focused benchmark scenarios: tracking after brief occlusions and tracking with multiple people and occlusions.

\begin{table*}[ht]
\centering
 {
      
\renewcommand{\arraystretch}{1.0} 
\setlength{\tabcolsep}{2.2pt}     
\resizebox{\textwidth}{!}{
\begin{tabular}{|l|l|l|l|l|l|l|l|l|l|l|l|p{1.5cm}|l|l|}
\hline
{Method} &
\multicolumn{8}{c|}{{All test set}} &
\multicolumn{3}{c|}{{Brief occlusions}} &
\multicolumn{3}{c|}{{Multiple people and occlusions}} \\
\cline{2-15}
& MOTA$\uparrow$ & IDF1$\uparrow$ & MT$\uparrow$ & ML$\downarrow$ & FN$\downarrow$ & FP$\downarrow$ & IDSW$\downarrow$ & FM$\downarrow$
& MOTA$\uparrow$ & IDF1$\uparrow$ & IDSW$\downarrow$
& MOTA$\uparrow$ & IDF1$\uparrow$ & IDSW$\downarrow$ \\
\hline
BoostTrack \cite{stanojevic2024boosttrack}
& 60.12 & 36.87 & 143 & 17 & 217654 & 89542 & 3801 & 7841
& 5.78 & 10.759 & \textbf{10}
& 9.15 & 16.56 & \textbf{2} \\
BoT\mbox{-}SORT \cite{aharon2022botsort}
& 56.52 & 35.33 & \textbf{245} & \textbf{6} & \textbf{102056} & 232112 & 4883 & 4991
& \textbf{12.42} & \textbf{19.15} & 91
& \textbf{21.831} & \textbf{33.88} & 10 \\
ByteTrack \cite{zhang2022bytetrack}
& 56.10 & 34.96 & 241 & \textbf{6} & 106270 & 230720 & 5400 & 5705
& 11.57 & 17.77 & 87
& 20.42 & 31.87 & 11 \\
OC-SORT \cite{cao2023ocsort}
& 62.86 & 34.55 & 208 & 9 & 137854 & 145336 & 6443 & 8434
& 10.71 & 17.10 & 51
& 15.50 & 25.89 & 6 \\
\hline
BoostTrack+$^\dagger$ \cite{stanojevic2024boosttrack} 
& 60.12 & 37.50 & 146 & 17 & 216848 & 90364 & 3823 & 7884
& 6.63 & 11.91 & 11
& 9.86 & 17.72 & \textbf{2} \\
BoostTrack++$^\dagger$ \cite{stanojevic2024btpp}
& 62.14 & 39.62 & 145 & 21 & 230102 & 62099 & 3069 & 7951
& 5.78 & 10.71 & 13
& 11.27 & 20 & \textbf{2} \\
BoT\mbox{-}SORT$^\dagger$ \cite{aharon2022botsort}
& \textbf{68.84} & \textbf{42.54} & 184 & 17 & 188089 & \textbf{52600} & \textbf{2298} & \textbf{4105}
& 9.86 & 17.01 & 36
& 15.49 & 25.89 & 6 \\
Deep\mbox{-}OC\mbox{-}SORT$^\dagger$ \cite{maggiolino2023deepocsort}
& 65.83 & 37.05 & 197 & 11 & 154621 & 106314 & 5575 & 8174
& 9.35 & 15.86 & 38
& 15.49 & 26.35 & 3 \\
StrongSORT$^\dagger$ \cite{du2023strongsort}
& 61.37 & 36.67 & 222 & 8 & 126838 & 166387 & 8054 & 8259
& 10.54 & 16.88 & 61
& 17.61 & 28.90 & 6 \\
\hline
\end{tabular}
}
}
\caption{{ Evaluation of different tracking methods on CHIRLA benchmark. 
%
Results on the full test set (8 sequences, 56 videos) are reported with MOTA, IDF1, MT/ML, FN, FP, IDSW and FM; the two focused scenarios (“Brief occlusions” and “Multiple people and occlusions”) report MOTA, IDF1 and IDSW. 
Arrows ($\uparrow$ / $\downarrow$) indicate whether higher or lower values are better. $^\dagger$ denotes models that incorporate appearance-embedding-based tracking. 
Appearance embeddings are extracted using a lightweight OSNet~\cite{zhou2019osnet} backbone (×0.25 scaling) pretrained on MSMT17 dataset.
For all methods, YOLOv11n \cite{yolo11_ultralytics_v11} is used as the person detector, and all trackers are executed with the default parameters from their official repositories.
}}
\label{tab:track-eval}
\end{table*}

For both scenarios, after running the trackers on the full test set, we selected the ground-truth frames corresponding to each scenario (as defined in Section Tracking Benchmark Data Generation). For each ground-truth ID, we retained the predicted bounding box with the highest IoU (>0.5), along with the corresponding tracker ID. Using this mapping, we then computed the evaluation metrics to assess whether trackers consistently maintained ID assignment before and after the occlusion conditions.

On the full test set, BoT-SORT and its appearance-augmented variant BoT-SORT$^\dagger$ \cite{aharon2022botsort} achieved the best overall results. BoT-SORT$^\dagger$ reached the highest accuracy, with MOTA 68.84 and IDF1 42.54, while also yielding the lowest FP (52,600), IDSW (2,298) and FM (4105). Meanwhile, BoT-SORT achieved the highest number of MT trajectories (245) and the lowest FP (102,056), and together with ByteTrack \cite{zhang2022bytetrack} recorded the smallest ML count (6). This confirms that trackers integrating appearance analysis provide a clear advantage in general conditions with continuous observations.

However, in the two focused scenarios: brief occlusions and multiple people with occlusions; appearance-based trackers underperform compared to traditional trajectory-based methods. For example, in the brief occlusion scenario, BoT-SORT achieved the best performance (MOTA 12.42, IDF1 19.15), whereas BoostTrack variants \cite{stanojevic2024boosttrack, stanojevic2024btpp} reduced IDSW to 10 but struggled in MOTA. Similarly, in the multiple people and occlusions scenario, BoT-SORT again led (MOTA 21.83, IDF1 33.88), while BoostTrack variants minimized IDSW (2) but at the cost of MOTA and IDF1.

We believe this inversion of performance reflects the challenge of Re-ID after occlusions. Appearance embeddings are sensitive to even small changes caused by lighting, pose, or occlusion, making it difficult for appearance-based trackers to maintain consistent IDs. In contrast, traditional trackers, which rely mainly on motion and trajectory continuity, are sometimes better able to reassign IDs after short occlusions, without taking into account appearance cues.



Overall, results in both scenarios dropped significantly compared to the full test set, from MOTA 68.84 / IDF1 42.54 to 12.42 / 19.15 in brief occlusions and 21.83 / 33.88 in multiple people and occlusions. This performance gap illustrates the challenging nature of CHIRLA’s occlusion-driven benchmarks and highlights the need for more robust strategies that can combine motion continuity with Re-ID embeddings tolerant to appearance changes.

In addition, Table \ref{tab:reid-eval} reports the results of several models across the four CHIRLA Re-ID scenarios. Performance is evaluated using CMC at rank-1, rank-5 and rank-10, together with mAP, on the test sets defined for each scenario (see Section Train and Test Data). We restrict the experiments to closed-set evaluation, where only known IDs are included in both train and test sets. Specifically, distractors (unknown IDs) provided by CHIRLA are excluded from the test set. We leave the open-set evaluation for future studies, as it requires models to define decision thresholds to distinguish between known and unknown IDs, which would enable a more realistic assessment of deployment conditions.

For all scenarios, we exclude the subsets \texttt{train\_0} and \texttt{test\_0} from evaluation, as these can be reserved for potential fine-tuning or model adjustment. Thus, the training set is used as the gallery and the test set as the query. Since the dataset is organized into multiple train–test subsets, we run experiments on each train–test pair and report the final results as the average across all subsets.

\begin{table}[!htb]
\centering
  {
      
\renewcommand{\arraystretch}{1.0} 
\setlength{\tabcolsep}{2.2pt} 
\resizebox{\textwidth}{!}
{
\begin{tabular}{|l|l|l|l|l|l|l|l|l|l|l|l|l|l|l|l|l|}
\hline
Method 
& \multicolumn{4}{c|}{Long term} 
& \multicolumn{4}{c|}{Multi-camera long term} 
& \multicolumn{4}{c|}{Multi-camera} 
& \multicolumn{4}{c|}{Reappearance} \\
\cline{2-17}
& \multicolumn{3}{c|}{CMC} & mAP 
& \multicolumn{3}{c|}{CMC} & mAP 
& \multicolumn{3}{c|}{CMC} & mAP 
& \multicolumn{3}{c|}{CMC} & mAP \\

\cline{2-4} \cline{6-8} \cline{10-12} \cline{14-16}
& @1 & @5 & @10 & 
& @1 & @5 & @10 & 
& @1 & @5 & @10 & 
& @1 & @5 & @10 & \\
\hline
ResNet50$^\ddagger$$^\circ$ \cite{He_2016_CVPR} & 13.30 & 33.19 & 49.39 & 18.22 & 18.02 & 36.92 & 61.58 & 30.65 & 42.58 & 60.02 & 75.00 & 45.39 & 50.98 & 75.30 & 89.92 & 54.98 \\
ResNet50$^\ddagger$$^x$\cite{He_2016_CVPR} & 11.64 & 33.47 & 49.57 & 16.96 & 16.55 & 37.51 & 59.18 & 28.50 & 36.21 & 57.02 & 69.97 & 40.02 & 52.04 & 76.91 & 89.35 & 54.36 \\
\hline
ResNet50-IBN$^\dagger$$^\circ$ \cite{pan2018IBN-Net, wang2018learning} & 4.54 & 22.04 & 40.36 & 12.84 & 11.12 & 37.30 & 63.93 & 23.30 & 13.24 & 45.21 & 64.21 & 25.82 & 19.61 & 50.39 & 74.90 & 31.36 \\
ResNet50-IBN$^\dagger$$^x$ \cite{pan2018IBN-Net, wang2018learning} & 8.48 & 33.05 & 50.26 & 15.18 & 16.47 & 48.56 & 74.53 & 29.04 & 21.83 & 48.09 & 65.17 & 28.70 & 19.52 & 58.46 & 80.04 & 31.24 \\
ResNet101-IBN$^\dagger$$^\circ$ \cite{pan2018IBN-Net, luo2019bag} & 15.85 & 39.18 & 57.15 & 20.16 & 20.49 & 41.88 & 67.80 & 33.51 & 38.41 & 60.54 & 75.26 & 42.79 & 57.99 & 82.31 & 93.07 & 60.38 \\
ResNet101-IBN$^\dagger$$^x$\cite{pan2018IBN-Net, luo2019bag} & 11.75 & 34.02 & 52.74 & 17.43 & 16.32 & 43.65 & 64.88 & 29.18 & 35.14 & 59.98 & 74.53 & 39.62 & 56.08 & 80.14 & 91.87 & 57.36 \\
ResNet101-IBN$^\dagger$$^\bullet$ \cite{pan2018IBN-Net, luo2019bag} & 12.22 & 32.81 & 51.14 & 18.45 & 22.48 & 52.63 & 73.93 & \textbf{35.28} & 53.45 & 72.79 & 83.45 & 53.25 & 61.00 & 83.29 & 93.00 & 61.15 \\
ResNet101-IBN$^\dagger$$^\circ$ \cite{pan2018IBN-Net, ye2021deep} & 12.61 & 36.48 & 56.19 & 19.64 & 19.88 & 43.33 & 68.68 & 32.08 & 39.43 & 59.15 & 80.30 & 43.97 & 58.37 & 82.12 & \textbf{94.38} & 60.59 \\
ResNet101-IBN$^\dagger$$^x$ \cite{pan2018IBN-Net, ye2021deep} & 12.94 & 34.41 & 55.24 & 17.96 & 17.58 & 46.34 & 71.53 & 29.92 & 38.94 & 67.21 & 80.39 & 43.16 & 55.85 & 81.32 & 90.49 & 55.20 \\
ResNet101-IBN$^\dagger$$^\bullet$ \cite{pan2018IBN-Net, ye2021deep} & 12.79 & 37.49 & 56.79 & 18.81 & 19.33 & 46.77 & 71.41 & 33.37 & 49.13 & 69.33 & 83.09 & 50.78 & \textbf{65.73} & 84.99 & 92.83 & 61.34 \\
ResNet101-IBN$^\dagger$$^\circ$ \cite{pan2018IBN-Net, he2023fastreid} & 17.47 & 44.17 & 63.85 & 22.68 & 18.85 & 43.86 & 69.73 & 32.79 & 45.52 & 66.73 & 80.31 & 48.06 & 61.30 & 83.02 & 93.02 & 61.55 \\
ResNet101-IBN$^\dagger$$^x$ \cite{pan2018IBN-Net, he2023fastreid} & 13.36 & 38.42 & 58.43 & 18.51 & 18.52 & 44.36 & 70.68 & 28.72 & 38.45 & 61.11 & 75.43 & 40.98 & 55.59 & 80.34 & 90.73 & 54.96 \\
ResNet101-IBN$^\dagger$$^\bullet$ \cite{pan2018IBN-Net, he2023fastreid} & 17.68 & 43.44 & \textbf{64.06} & 21.46 & 20.02 & 49.48 & 74.18 & 30.53 & 54.54 & \textbf{73.24} & 85.58 & 52.61 & 61.29 & 83.93 & 92.91 & 59.39 \\
\hline
ResNet50* \cite{He_2016_CVPR, wang2018learning} & 15.74 & 40.90 & 60.60 & 20.28 & 21.10 & 47.01 & 72.36 & 32.28 & 50.00 & 67.35 & 77.11 & 49.72 & 54.33 & 81.28 & 89.92 & 56.64 \\
ResNet101* \cite{He_2016_CVPR, wang2018learning} & 18.58 & 43.76 & 62.07 & 22.01 & \textbf{24.31} & 45.79 & 67.18 & 33.93 & 50.70 & 68.33 & 77.50 & 49.97 & 56.88 & 82.01 & 92.20 & 58.69 \\
ResNet152* \cite{He_2016_CVPR, wang2018learning} & 17.38 & 42.42 & 63.17 & 22.03 & 20.09 & 42.63 & 74.35 & 32.69 & 51.53 & 69.80 & 78.63 & 50.82 & 58.30 & 83.92 & 92.25 & 60.09 \\
ResNet50-IBN* \cite{pan2018IBN-Net, wang2018learning} & 8.61 & 26.10 & 43.39 & 15.79 & 14.40 & 38.81 & 58.15 & 27.78 & 23.61 & 47.67 & 68.64 & 33.78 & 43.64 & 65.90 & 84.57 & 46.33 \\
ResNet101-IBN* \cite{pan2018IBN-Net, wang2018learning} & 7.57 & 23.30 & 39.54 & 15.43 & 14.78 & 40.31 & 62.55 & 28.01 & 27.58 & 47.53 & 70.73 & 36.65 & 39.26 & 64.06 & 81.18 & 44.28 \\
ResNet152-IBN* \cite{pan2018IBN-Net, wang2018learning} & 7.68 & 25.22 & 39.58 & 15.41 & 18.14 & 40.23 & 62.89 & 30.24 & 22.96 & 45.00 & 68.00 & 34.16 & 41.47 & 67.58 & 84.28 & 46.94 \\
\hline
ConvNext-B \cite{Liu_2022_CVPR} & 4.98 & 21.79 & 39.72 & 12.42 & 14.36 & 43.82 & 72.41 & 28.14 & 11.96 & 35.88 & 67.50 & 23.80 & 20.07 & 55.90 & 76.36 & 32.97 \\
ConvNext-S \cite{Liu_2022_CVPR} & 11.48 & 38.90 & 59.62 & 16.26 & 21.85 & 47.68 & 69.75 & 30.77 & 14.55 & 39.57 & 67.11 & 25.01 & 17.97 & 50.53 & 71.47 & 30.06 \\
ConvNext-T \cite{Liu_2022_CVPR} & 7.23 & 30.77 & 49.05 & 13.41 & 14.35 & 46.30 & 72.37 & 27.35 & 14.61 & 44.68 & 66.51 & 22.40 & 17.34 & 49.89 & 74.00 & 32.88 \\
EdgeNext-B \cite{maaz2022edgenext} & 5.85 & 26.58 & 52.43 & 14.63 & 7.90 & 39.90 & 63.58 & 23.84 & 13.50 & 45.80 & 63.43 & 26.72 & 17.52 & 50.69 & 76.45 & 29.35 \\
EdgeNext-S \cite{maaz2022edgenext} & 7.31 & 31.10 & 53.12 & 15.26 & 10.95 & 32.64 & 57.51 & 23.92 & 14.64 & 51.69 & 78.38 & 25.79 & 20.31 & 50.82 & 75.26 & 31.75 \\
EdgeNext-XS \cite{maaz2022edgenext} & 9.11 & 28.54 & 39.94 & 14.60 & 17.33 & 50.35 & 70.46 & 29.89 & 12.24 & 47.90 & 64.40 & 24.34 & 17.17 & 46.17 & 77.39 & 30.59 \\
FastViT-S12 \cite{vasu2023fastvit} & 7.99 & 27.85 & 53.15 & 15.45 & 13.00 & 42.04 & 71.73 & 26.57 & 12.85 & 41.00 & 59.95 & 23.86 & 26.17 & 51.69 & 74.02 & 33.81 \\
FastViT-SA12 \cite{vasu2023fastvit} & 8.56 & 21.50 & 40.14 & 13.02 & 9.61 & 37.20 & 52.12 & 23.10 & 20.01 & 48.93 & 70.65 & 29.78 & 25.12 & 58.47 & 84.79 & 32.38 \\
FastViT-SA24 \cite{vasu2023fastvit} & 8.48 & 20.60 & 38.78 & 12.50 & 7.26 & 45.41 & 75.18 & 24.07 & 19.48 & 44.83 & 67.22 & 28.44 & 19.31 & 47.05 & 74.23 & 30.48 \\
GhostNet 0.5$\times$ \cite{ghostnet} & 14.62 & 38.89 & 58.22 & 20.80 & 20.16 & 45.63 & 72.89 & 33.30 & 50.31 & 67.73 & 78.20 & 50.42 & 53.91 & 80.60 & 90.88 & 56.75 \\
GhostNet 1.0$\times$ \cite{ghostnet} & 15.88 & 39.86 & 61.16 & 22.20 & 17.94 & 42.50 & 74.79 & 31.21 & 51.64 & 69.99 & 81.56 & 52.25 & 59.27 & 83.44 & 93.00 & 61.40 \\
GhostNet 1.3$\times$ \cite{ghostnet} & 16.03 & 40.01 & 61.35 & 22.37 & 16.72 & 46.20 & 78.00 & 32.23 & 51.73 & 70.17 & 80.61 & 53.35 & 61.58 & 84.91 & 92.78 & 62.77 \\
RepViT-M0.9 \cite{wang2024repvit} & 3.88 & 36.87 & 55.48 & 13.62 & 8.39 & 52.29 & 74.03 & 23.96 & 20.30 & 53.57 & 74.96 & 26.21 & 17.35 & 61.88 & 81.19 & 28.12 \\
RepViT-M1.0 \cite{wang2024repvit} & 7.22 & 31.38 & 60.07 & 13.55 & 10.26 & \textbf{60.97} & 85.51 & 23.46 & 16.36 & 56.48 & 86.48 & 25.30 & 22.85 & 70.39 & 88.45 & 29.55 \\
RepViT-M1.5 \cite{wang2024repvit} & 7.46 & 33.93 & 48.08 & 13.40 & 16.69 & 46.54 & \textbf{87.65} & 25.60 & 10.72 & 53.37 & \textbf{86.49} & 21.91 & 18.32 & 68.29 & 83.71 & 27.49 \\
ResNet50 \cite{He_2016_CVPR} & 14.23 & 39.28 & 59.72 & 20.20 & 17.74 & 43.32 & 72.81 & 31.27 & 52.16 & 70.64 & 79.71 & 53.31 & 55.06 & 81.71 & 91.49 & 58.57 \\
ResNet101 \cite{He_2016_CVPR} & \textbf{18.81} & \textbf{44.27} & 63.25 & \textbf{23.24} & 19.32 & 47.82 & 69.99 & 33.24 & 53.46 & 70.02 & 78.37 & 54.26 & 62.55 & 85.53 & 93.04 & 62.88 \\
ResNet152 \cite{He_2016_CVPR} & 18.21 & 43.98 & 62.84 & 23.04 & 21.11 & 44.33 & 75.55 & 33.85 & \textbf{56.01} & 72.35 & 81.10 & \textbf{55.90} & 63.59 & \textbf{86.22} & 93.03 & \textbf{63.85} \\
ResNet18-IBN \cite{pan2018IBN-Net} & 9.41 & 25.64 & 40.73 & 16.25 & 13.91 & 38.75 & 61.93 & 26.91 & 20.95 & 47.54 & 66.69 & 31.97 & 30.89 & 59.35 & 83.32 & 40.71 \\
ResNet50-IBN \cite{pan2018IBN-Net} & 8.09 & 27.09 & 43.70 & 15.67 & 15.38 & 39.90 & 57.85 & 27.86 & 21.99 & 46.54 & 68.36 & 32.61 & 43.09 & 65.79 & 83.19 & 45.86 \\
ResNet101-IBN \cite{pan2018IBN-Net} & 8.54 & 24.16 & 39.68 & 15.71 & 16.05 & 40.79 & 62.88 & 28.91 & 26.68 & 49.20 & 69.69 & 36.45 & 38.87 & 62.46 & 80.75 & 44.03 \\
ResNet152-IBN \cite{pan2018IBN-Net} & 7.74 & 24.99 & 39.15 & 15.37 & 17.59 & 42.32 & 64.84 & 30.32 & 23.60 & 45.28 & 67.16 & 34.12 & 40.46 & 68.05 & 84.42 & 46.78 \\
Swin-S \cite{liu2021swin} & 7.02 & 25.81 & 52.77 & 14.88 & 16.18 & 51.18 & 74.66 & 27.51 & 19.77 & 50.90 & 73.58 & 29.49 & 16.94 & 47.38 & 75.47 & 30.40 \\
Swin-T \cite{liu2021swin} & 11.22 & 43.75 & 61.95 & 15.20 & 11.74 & 46.62 & 76.67 & 25.71 & 14.04 & 41.01 & 67.33 & 27.15 & 16.88 & 51.21 & 77.45 & 30.98 \\
ViT-S \cite{dosovitskiy2020vit} & 6.11 & 27.12 & 43.88 & 14.73 & 15.84 & 40.77 & 58.55 & 27.17 & 15.93 & 38.44 & 66.55 & 29.76 & 20.15 & 42.20 & 66.64 & 32.39 \\
ViT-T \cite{dosovitskiy2020vit} & 7.14 & 29.19 & 45.85 & 15.05 & 16.31 & 38.08 & 67.90 & 26.59 & 14.63 & 42.32 & 63.73 & 29.94 & 19.44 & 40.69 & 66.40 & 32.38 \\
VOLO-D1 \cite{yuan2022volo} & 12.66 & 34.09 & 53.90 & 17.95 & 19.50 & 42.41 & 70.24 & 31.80 & 38.57 & 58.23 & 71.18 & 40.95 & 45.59 & 73.17 & 88.45 & 49.98 \\
VOLO-D2 \cite{yuan2022volo} & 12.35 & 34.50 & 50.98 & 17.59 & 17.46 & 44.24 & 69.91 & 31.10 & 40.02 & 59.66 & 70.94 & 41.78 & 49.70 & 75.03 & 85.68 & 49.97 \\
VOLO-D3 \cite{yuan2022volo} & 13.99 & 36.84 & 54.40 & 18.94 & 20.36 & 48.81 & 70.59 & 30.41 & 35.33 & 64.93 & 77.21 & 40.99 & 53.83 & 81.97 & 92.54 & 54.11 \\
\hline
\end{tabular}
}
}
\caption{{ 
Evaluation of the four person re-identification scenarios defined in our proposed benchmark using different methods in closed-set evaluation.  
$^\circ$ indicates specific pre-training on Market1501, $^x$ on DukeMTMC, and $^\bullet$ on MSMT17  datasets. 
\textit{*} denotes results with an input size of $384\times128$, while all other results use $256\times128$. 
Model weights for $^\ddagger$ are extracted from CentroidsReid \cite{wieczorek2021unreasonable}, 
for $^\dagger$ from FastReid \cite{he2023fastreid}, 
and the remaining models from ReIDZoo \cite{zuo2024cross}. }
}
\label{tab:reid-eval}
\end{table}

The long-term scenario emerges as the most challenging, as it requires consistent Re-ID  over extended sequences with large temporal gaps. This is reflected in the lowest CMC values among all scenarios. The best model here is ResNet101 \cite{He_2016_CVPR} (256×128 input size), which reached 18.81\% CMC@1 and 23.24\% mAP, highlighting the difficulty of maintaining reliable embeddings across long time spans.

The multi-camera long-term scenario results as the second most difficult. In this case, a ResNet101 backbone with a larger input size of 384×128 achieved the best CMC@1 result (24.31\%), while ResNet101-IBN$^\dagger$ \cite{pan2018IBN-Net} obtained the best mAP (35.28\%). 

In the multi-camera scenario (without the long-term constraint), performance improves considerably. Here, ResNet152 \cite{He_2016_CVPR} achieved the highest scores, with 56.01\% CMC@1 and 55.90\% mAP, demonstrating the advantage of deeper backbones when IDs are observed across multiple views without long interruptions.
Although this setting remains challenging due to differences in camera angles and positions, IDs still share more appearance consistency than in the long-term scenarios, which makes this scenario comparatively more feasible.

The reappearance scenario, in contrast, proved the most manageable. ResNet101-IBN$^\dagger$ achieved the best CMC@1 (65.73\%), while ResNet152 obtained the highest mAP (63.85\%). These results suggest that reappearances, often within the same camera view, are handled relatively well by current methods compared with the other analyzed scenarios.

Overall, ResNet-based models showed the most consistent and robust performance across all CHIRLA scenarios. While reappearance is comparatively easier, the long-term and multi-camera long-term scenarios remain highly challenging and highlight the need for models that are more resilient to temporal gaps and inter-camera variations.

}

\section*{Usage Notes}

The CHIRLA dataset provides a comprehensive resource for the development and evaluation of person Re-ID algorithms in challenging, long-term scenarios. It enables testing the capability of models to identify persons across different cameras, times, and conditions, including variations in clothing and appearance.
Although Re-ID is the primary focus, the dataset can also be used for tracking algorithms. It allows for the evaluation of an algorithm's ability to detect, identify, and maintain consistent IDs within single-camera settings. Additionally, the dataset supports scenarios such as tracking through occlusions and multi-person interactions.

The dataset is publicly available and data collection adhered to ethical guidelines, with informed consent obtained from all participants.

\section*{Code Availability}
\label{sec:code-availability}


The information for downloading the dataset and the software to be used for the evaluation of the systems on this dataset can be found in the following GitHub repository: \url{https://github.com/bdager/CHIRLA}.

\bibliography{sample}


\section*{Acknowledgements}


This paper is part of the grant PID2022-138453OB-I00 funded by MICIU/AEI/10.13039/501100011033 and by “ERDF A way of making Europe”. Bessie Dominguez-Dager has been funded by the grant  FPU23/02247 from Ministerio de Ciencia, Innovacion y Universidades of Spain.

\section*{Author contributions statement}


B. D.: Software, Validation, Methodology, Data Curation, Writing. F. E.: Software, Formal analysis, Data Curation, Methodology. F. G.: Conceptualization, Writing - Original Draft, Supervision. M. C.: Resources, Writing - Review \& Editing, Funding acquisition.

\section*{Competing interests} 


The authors declare no known competing financial interests.






\end{document}